\documentclass[]{spie}  %>>> use for US letter paper
\usepackage{amsmath,amsfonts,amssymb}
\usepackage{graphicx}
\usepackage[colorlinks=true, allcolors=blue]{hyperref}
\usepackage{subcaption}
\usepackage{xcolor}

\title{Enhancing the quality of gauge images captured in smoke and haze scenes through deep learning}
\author[a]{Oscar H. Ram\'irez-Agudelo}
\author[a]{Akshay N. Shewatkar}
\author[a,b]{Edoardo Milana}
\author[c]{Roland C. Aydin}
\author[a]{Kai Franke}
\affil[a]{German Aerospace Center (DLR), Institute for the Protection of Terrestrial Infrastructures, Rathausallee 12, 53757 Sankt Augustin, Germany}
\affil[b]{Department of Microsystems Engineering (IMTEK), University of Freiburg, Georges-K\"ohler-Allee 103, 79110 Freiburg, Germany}
\affil[c]{Helmholtz Center Hereon, Institute of Material Systems Modeling, Max-Planck-St 1, 21502 Geesthacht, Germany}

\authorinfo{Further author information: (Send correspondence to O.H.R-A.)\\O.H.R-A.: E-mail: Oscar.RamirezAgudelo@dlr.de%, Telephone: +49 2241 20148 35
}

\begin{document} 
\maketitle

\begin{abstract}
Images captured in hazy and smoky environments suffer from reduced visibility, posing a challenge when monitoring infrastructures and hindering emergency services during critical situations. The proposed work investigates the use of the deep learning models to enhance the automatic, machine-based readability of gauge in smoky environments, with accurate gauge data interpretation serving as a valuable tool for first responders. The study utilizes two deep learning architectures, FFA-Net and AECR-Net, to improve the visibility of gauge images, corrupted with light up to dense haze and smoke. Since benchmark datasets of analog gauge images are unavailable, a new synthetic dataset, containing over 14,000 images, was generated using the Unreal Engine. The models were trained with an 80\% train, 10\% validation, and 10\% test split for the haze and smoke dataset, respectively. For the synthetic haze dataset, the SSIM and PSNR metrics are about 0.98 and 43\,dB, respectively, comparing well to state-of-the art results. Additionally, more robust results are retrieved from the AECR-Net, when compared to the FFA-Net. Although the results from the synthetic smoke dataset are poorer, the trained  models achieve interesting results. In general, imaging in the presence of smoke are more difficult to enhance given the inhomogeneity and high density. Secondly, FFA-Net and AECR-Net are implemented to dehaze and not to desmoke images. This work shows that use of deep learning architectures can improve the quality of analog gauge images captured in smoke and haze scenes immensely. Finally, the enhanced output images can be successfully post-processed for automatic autonomous reading of gauges.
\end{abstract}


\section{Introduction} \label{sec:intro}  
% Section in construction
Haze and smoke affect the perception of computer vision tasks. These atmospheric phenomena scatters and attenuates the light due to the presence of dust and particles. The degradation in the quality of the objects appearance and contrast has an important impact for post-processing tasks. In cases of strong presence of haze or smoke, it can affect the image quality severely. This might pose a massive problem for surveillance critical infrastructures. In the event of a disaster, or even an attack, haze and smoke might hinder emergency services such as fire-fighters. Therefore, it is important to develop technologies to help improve the safety and security of critical infrastructures.\newline 

In the literature, the concept of removing haze from images, \textit{a.k.a.} dehazing, has been exploited. The approaches to address this are based on prior-based methods as well as the implementation of convolutional neural networks (CNNs; i.e. deep learning methods) in order to improve the image vision (see Sect.~\ref{sec:sota}). The concept of smoke-removal has not yet been widely studied in depth due to its intrinsic complexity (see Sects.~\ref{sec:sota} and ~\ref{sec:discussion}). In the recent past, deep learning frameworks have received special attention due to its ability to dehaze images and superiority in performance with respect prior-based methods. However, it is well known and accepted the requirement of large and relevant quantity of data to acquire a robust trained model.\newline

CNNs dehazing models often rely on synthetic hazy training sets. The generation of synthetic datasets has offered a useful alternative to bridge the gap of the scarcity of suitable sets \cite{zheng20234khaze}. For instance, RESIDE \cite{li2019reside} is the benchmark dataset of both synthetic and real-world hazy images containing indoor and outdoors scene (see Sects.~\ref{sec:sota} and ~\ref{sec:discussion}). RESIDE has been used by different deep learning dehazing frameworks \cite{liu2019griddehazenet,qin2019ffanet,wu2021contrastive}. This work focuses in analog gauges as many industrial environments rely on analog gauges to monitor process and conditions of infrastructure. Due to their tasks, such environments are susceptible to scenarios involving the decrease of integrity due to fire. To have a model ready at hand to remove the haze and smoke of gauges reader images would be beneficial.\newline 

The main goal of this work is to create train models to enhance the image quality of analog gauges under the presence of haze and smoke. Therefore, first, it is created a hazy and also smoky datasets of analog gauges and then proceed to use different methods to remove the haze and smoke, respectively. This work is structured as follows. Section~\ref{sec:sota} reviews relevant literature. Section~\ref{sec:method} describes the methodology adopted in this work. The results are presented in Section~\ref{sec:results} and discussed in Section~\ref{sec:discussion}. The conclusion is presented in  Section~\ref{sec:conclusion}.

\section{Related works}\label{sec:sota}
\subsection{Dehazing}

\subsubsection{Classical approaches}\label{sec:classic_approaches}

The standard approaches (non-deep learning approaches) are prior-based methods. The most popular are: the image dehazing using a \textit{Dark Channel Prior} (DCP) \cite{he2009dcp} and \textit{Efficient Image Fehazing with Boundary Constraint and Contextual Regularization} (BCCR) \cite{meng2013bccr}. In DCP, the authors based their method on assuming that most local patches in haze-free outdoor images contain some pixels which have very low intensities in at least one-color channel. With this assumption the scene radiance and transmission are obtained. BCCR is an optimization method aimed to, first, determine the transmission function by assuming a radiance cube with boundary constraint. The above, combined with a weighted $L_{1}-$norm based contextual regularization, is modeled into an optimization problem to estimate the unknown scene transmission.\newline
 
The aforementioned prior-based methods aim to model the atmosphere scattering model proposed by Cartney et al. in 1976 \cite{McCartney_1976}. The equation is as follows:

\begin{eqnarray}
\mathbf{I}(x) & =&  t(x) \mathbf{J}(x) + (1 - t(x))\mathbf{A} \label{eq:1} \\
t(x)          & = & \exp^{\left[-\beta\,d(x)\right]} 
\label{eq:2}
\end{eqnarray}

where $\mathbf{I}(x)$ is the observed image, $\mathbf{J}(x)$ is the scene radiance, $\mathbf{A}$ is the global atmospheric light, and $t(x)$ is the scene transmission (value between 0 and 1), and $\beta$ is the medium extinction or scattering coefficient of atmosphere. This problem described by Eqs.~\ref{eq:1} and~\ref{eq:2} has more variables than equations that could be modeled; i.e. they form an \textit{ill-posed} problem. Therefore, the different approaches try to solve this under constrained problem.

\subsubsection{Deep learning}\label{sec:ffanet}

Over the past years, there has been a large effort spent on employing Convolutional Neuronal Networks (CNNs) to remove the haze from RGB images. While they do not rely on any prior knowledge, their main requirement is to have sufficient data to train a model, with the quality of the model in general improving until a certain point with the amount of available data. Thus, it is mandatory to have a relevant volume of data, depending on the application, to achieve a sufficiently good model performance. Some of the existing and known deep learning based dehazing methods are GCANet from Chen et al.\cite{chen2018gated} and DehazeNet from Cai et al. \cite{cai2016dehazenet}. GCAnet avoids gridding artifacts by proposing a smoothed dilated convolution. Afterwards, a gated subnetwork is applied to fuse the features of different levels. DehazeNet estimates the medium transmission map by implementing a deep CNN. The transmission map is subsequently used to recover a haze-free image via atmospheric scattering model \cite{McCartney_1976} (see Eq.~\ref{eq:1}.\newline

Other interesting work is the \textit{feature fusion attention network for single image dehazing} (FFA-Net) \cite{qin2019ffanet}. Given its versatility, the method itself is an improvement compared to classical approaches. The feature fusion attention (FFA) network is a novel method for image dehazing, where its 'end-to-end' CNN learns the mapping from a hazy image to a clear image in a supervised learning regime. The feature fusion network (FFA-Net) is made up of three components \cite{qin2019ffanet}: \textbf{i)} A novel feature attention ($FA$) module consisting of channel attention ($CA$) and pixel attention ($PA$) layers (see Fig.~3 of Qin et al. \cite{qin2019ffanet}). \textbf{ii)} Basic blocks, where each basic block consists of an $FA$-module and local residual learning, and \textbf{iii)} an attention based feature fusion structure. It can retain shallow layers information and pass it into deep layers. It can fuse the information from the different features and also adaptively learn the weights of the different level features (for more details we refer the reader to the FFA-Net manuscript \cite{qin2019ffanet}).

\subsubsection{Contrastive Representation Learning}\label{sec:aecrnet}
Contrastive learning has recently gained attention for its versatility in self-supervised as well as supervised representation learning. This is done via the implementation of the \textit{contrastive loss}. For a given anchor point, contrastive learning aims to pull the anchor close to the positive samples and to push the anchor away from the negative samples in the representation space.\newline

The paper \textit{Contrastive Learning for Compact Single Image dehazing} (AECR-Net) \cite{wu2021contrastive} proposed a contrastive regularization based on contrastive learning. It is an end-to-end deep learning dehazing network which makes use of autoencoders. The basic block of the autoencoder-like network is based on the FFA-Net, where the FA block becomes the (basic) block in the proposed structure of the network. It also has an adaptive mixup operation and dynamic feature enhancement module (see Fig.~3 of Wu et al.\cite{wu2021contrastive} of section 3 and references therein). At this point, AECR-Net surpassed the state-of-the-art approaches, obtaining a high PSNR score (see Sect.~\ref{sec:psnr}) by also greatly reducing the number of parameters to fit (see also their Fig~2).\newline

\subsubsection{Vision Transformers}\label{sec:transformers}
A transformer is a component used in neuronal network design for processing sequential data, such as natural language data, time series data, etc. It has an encoder-decoder structure which includes a stack of N=6 identical layers, multi-head attention mechanism, residual connection and a layer of normalisation (for details see Section 3.1 and Figure 1 of Vaswani et al.\cite{vaswani2017attention}). Its architecture is based on attention mechanisms, which can be described as mapping a query and a set of key-value pairs to an output (see Equation 1 in Section 3.2 of Vaswani et al.\cite{vaswani2017attention}). Compared to recurrent neuronal networks (RNNs) such as long short-term memory (LSTM), the implementation is more versatile as it requires less training time. 

Song et al.\cite{song2023dehazeformer} makes use of vision transformers for the purpose of single image dehazing. This research is a state-of-the-art approach based on Swin Transformers\cite{liu2021swintransformer}. Their authors proposed a hierarchical computer vision transformer whose representation is computed with shifted windows. 
The implementation proved to be beneficial for multi-layer-perceptron architectures and improved the performance for object detection and semantic segmentation for benchmark datasets such as COCO. For more details, we refer the reader to Figs.~1~to~3 of Liu et al.\cite{liu2021swintransformer}.\newline

\subsection{Desmoking}\label{sec:desmoking}
%To explain few more works
The authors in Chen et al.\cite{chen2022desmoke} implemented \textit{DesmokeNet}, a novel two-stage deep learning architecture to remove smoke from images. In the first stage, the \textit{smoke removal network} (SRN) aims to remove the light and thick smoke locally from the image. The second stage, the \textit{pixel compensation network} (PCN) recovers the missing pixels. The SRN has with it four modules; namely: (i) a smoke mask estimation module (SME), (ii) removing smoke modules (RMS), (iii) enhancers, and (iv) a by-pass feature sharing module (BFS). The SME predicts the global smoke mask. The three RMS modules removes the light and thick smoke by adopting pyramid-style connections to extract the low- and high-level features. The \textit{enhancers} aim to improve the image quality by means of multi-scale CNNs. The BFS shares the information between SME and RMS. The reason is that both will benefit from a better estimation of the smoke mask and the smoke free image. This stage already produces a good result named ``coarse output''.\newline

The PCN recovers the information of the image for the pixels where there may still remain either residual smoke or missing content. It consists of two modules: (i) thick smoke mask generation and (ii) residual estimation. The former is done by constructing a smoke map. Then, the residual estimation is composed of an encoder-decoder architecture with skip connections. The outcome of this network is known in the paper as ``fine output''.  %Finally, explain Loss

\section{Methodology}\label{sec:method}

This section describes the methodology adopted to recover a corrupted image from haze and smoke. First, the steps to take into account to why it was necessary to acquire data of gauges in Sect.~\ref{sec:dataset_requirements}, where after the gauge dataset are introduced in Sect.~\ref{sec:datasets}. The adaptation implemented to the three methods selected in this work are explained in Sect.~\ref{sec:implementation_of_methods}. Finally, the metrics used to evaluate the models are presented in Sect.~\ref{sec:metrics}.

\subsection{Dataset}\label{sec:dataset_requirements}
Datasets are essential for training any machine learning model. The more data a model has, the better its ability to train.  %and hence enhance the predictions it can make 
In machine/deep learning there are different synthetic datasets available to train indoor and outdoor models to dehaze images; such as: Fattal's \cite{fattal2008},  D-Haze \cite{ancuti2016dhazy}, Haze RD , Dense-Haze \cite{codutra2019densehaze}, RESIDE \cite{li2019reside} datasets. The dataset should fulfill the requirement of mapping a single clear image to different hazy images, each containing different density of haze. For instance, the RESIDE dataset for every ground truth it has ten different hazy images of increasing density. This enables that the network is trained with good enough variation and diversity of cases and therefore can be trained efficiently.

In this paper, we aim to reconstruct images of gauges. In the literature, however, there is no single dataset available for this task. Therefore, it was necessary to create a diverse dataset containing both, clear, sharp RGB images of gauges as well as respective hazy/smoky images.\newline

\subsubsection{Synthetic dataset $-$ Unreal Engine}\label{sec:ue}

%%%%%%%%%%%% KAI %%%%

The capturing of authentic, real-world images of gauges in haze- or smoke-filled environments is challenging. On the one hand, enough images must be acquired to generate a sufficiently large data set. On the other hand, it is difficult to generate quantifiable situations from the combination of lighting conditions, environments, gauges, and haze and smoke properties in reality. In particular, the density of haze or smoke is difficult to reproduce, which consequently complicates the evaluation of the effectiveness of the image enhancement presented in this paper.\newline

In order to improve the visibility of gauge images taken in a haze- and smoke-filled environment, the integration of a synthetic dataset is included. The generation of these synthetic images poses some challenges. 
To facilitate the generation of images that mimic real-life haze- and smoke-corrupted photos of gauges, properties of the environment and haze/smoke need to be respected. This includes realistic rendering of the environment as well as the gauges themselves and the integration of haze/smoke that reduces visibility as depth increases. Lighting in general needs to be modeled as realistic as possible, since it may affect the visibility of the gauge in positive (e.g. by providing more brightness to an otherwise dark image) and negative (e.g. scattering or reflections of light on the gauge's display) ways.\newline

On the other hand, the modeled synthetic environment needs to be variable with regard to environment, gauges and haze/smoke. This includes the modification of lighting settings, gauges, smoke and the surrounding environment in general. In addition, the position of the virtual camera should be changed to provide different perspectives on the gauge.

In order to navigate these hurdles, a game engine is employed. Game engines are proficient in simulating real-life dynamics to a certain extent. This includes the ability to provide realistic rendering in virtual environments. They also are very versatile, being able to display virtual environments from different perspectives and using different environment and gauge models. Game engines are already used in similar applications to the one presented in this paper \cite{2019rosiqueSystematic}, justifying the use of game engines to create synthetic images.

\begin{figure}
\subfloat[Resulting image of a gauge with a reflective cover glass (excluding smoke).]{
	\includegraphics[width=0.5\textwidth]{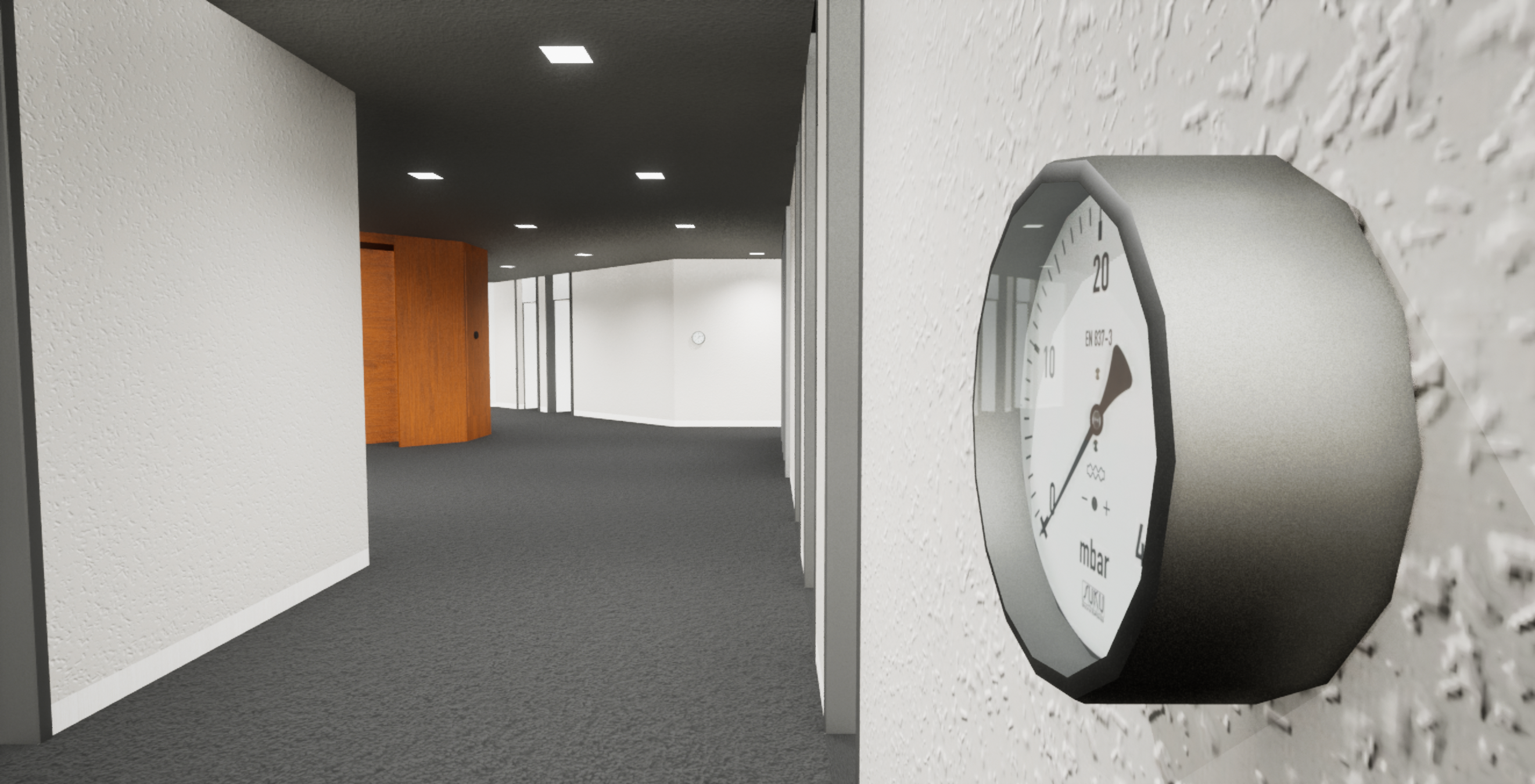}
}\\
\subfloat{
	\subfloat[Base color.]{
		\includegraphics[width=0.22\textwidth]{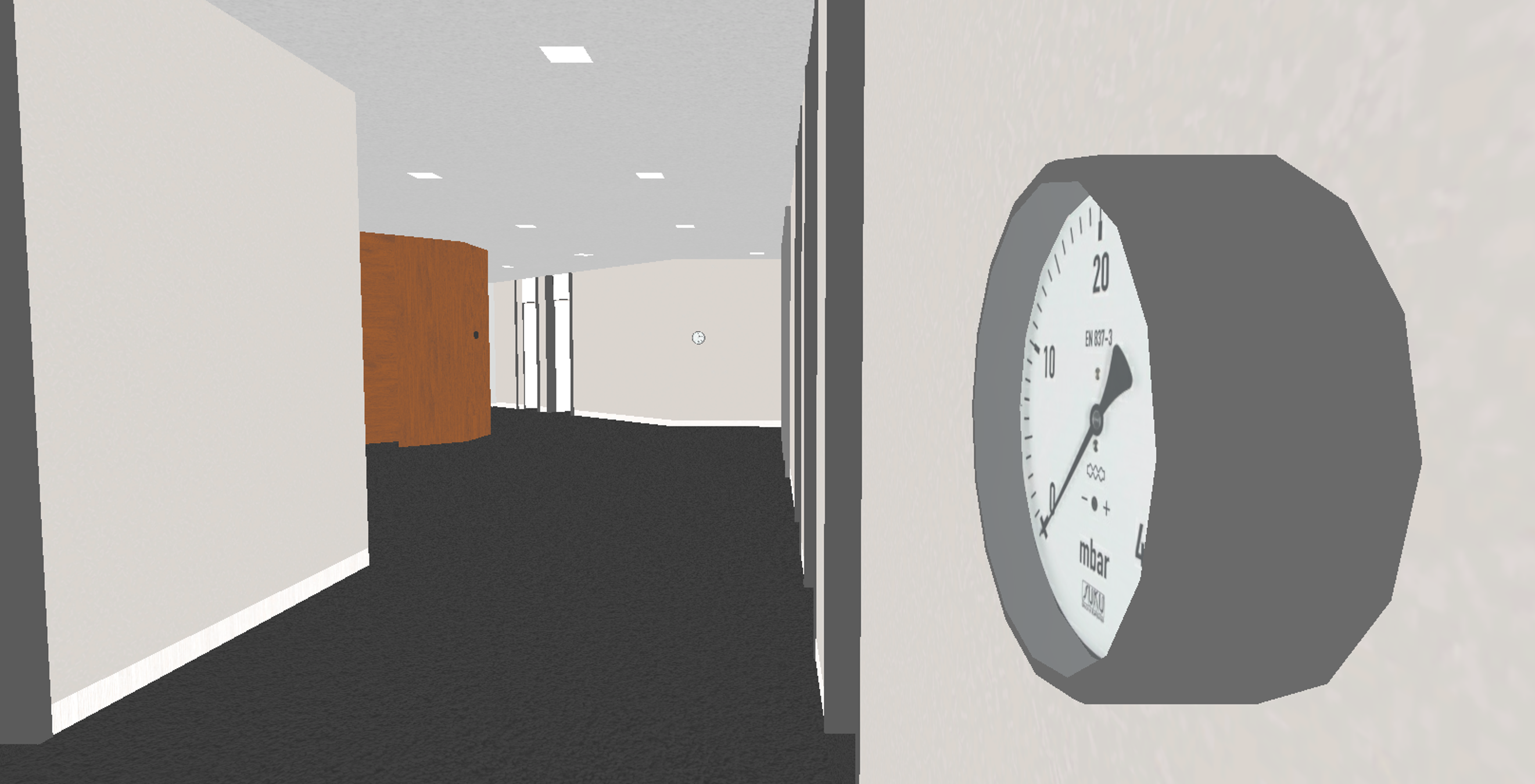}
	}
	\subfloat[Lighting influence.]{
		\includegraphics[width=0.22\textwidth]{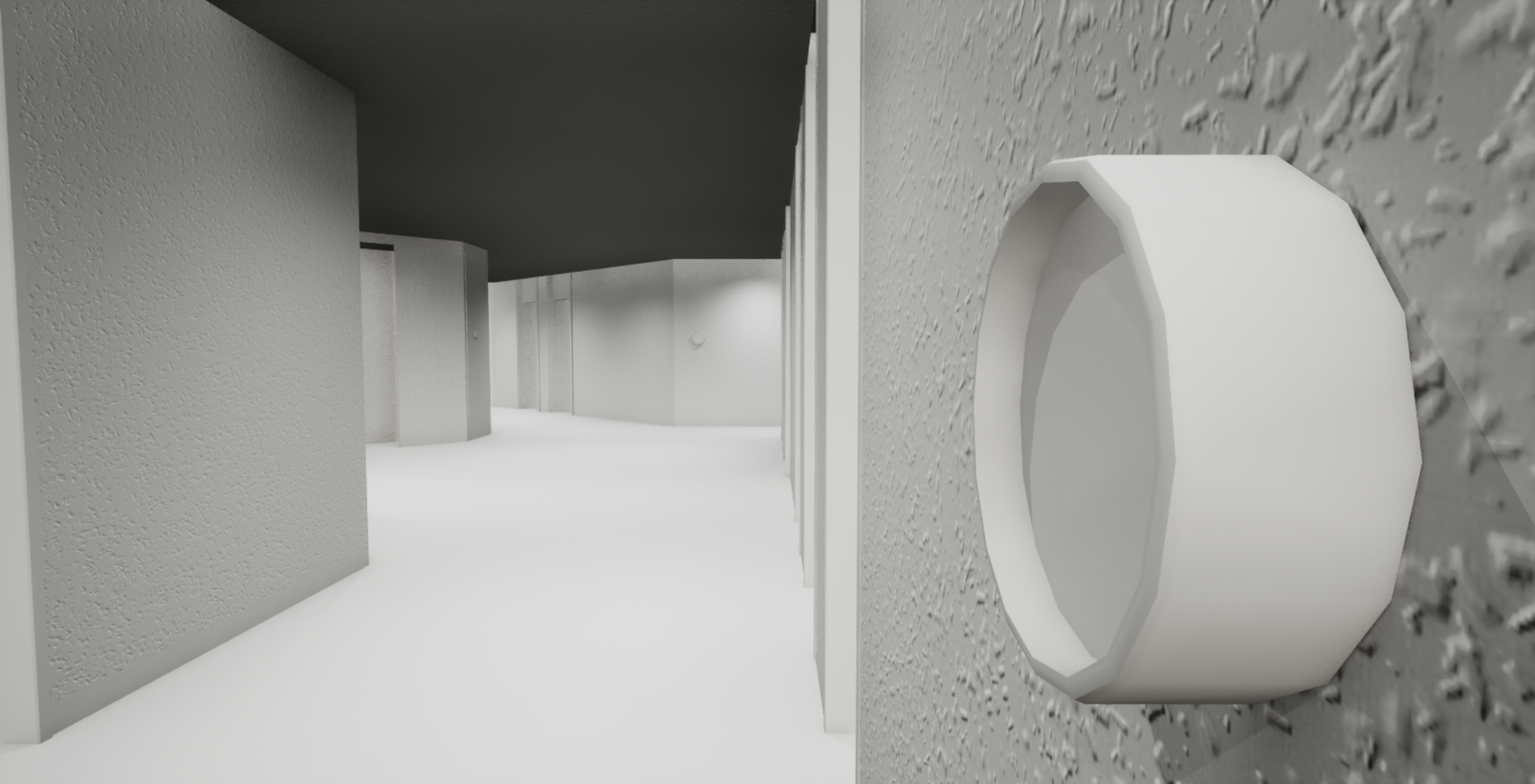}
	}\\
	\subfloat[Depth perception.]{
		\includegraphics[width=0.22\textwidth]{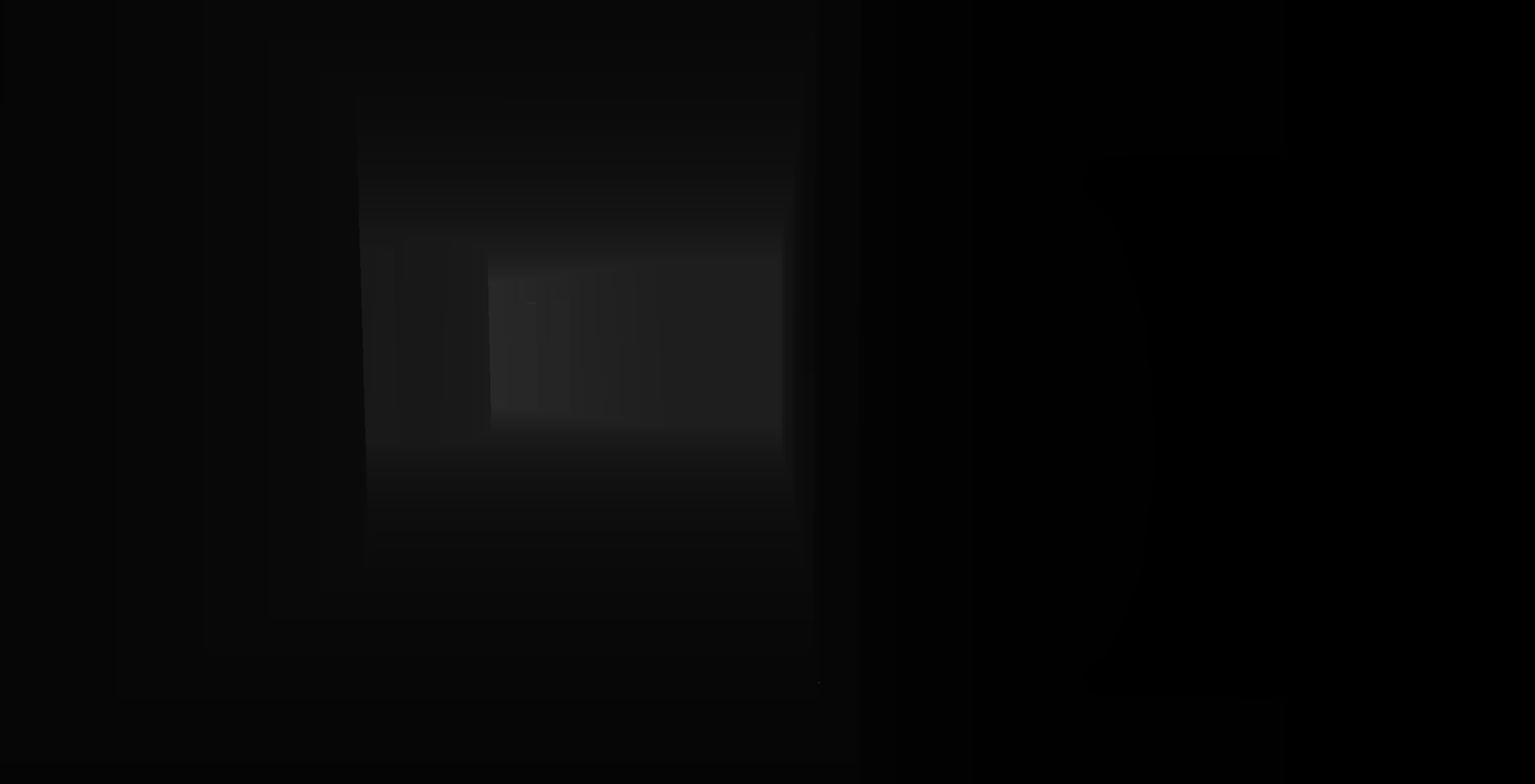}
	}
	\subfloat[Surface reflection.]{
		\includegraphics[width=0.22\textwidth]{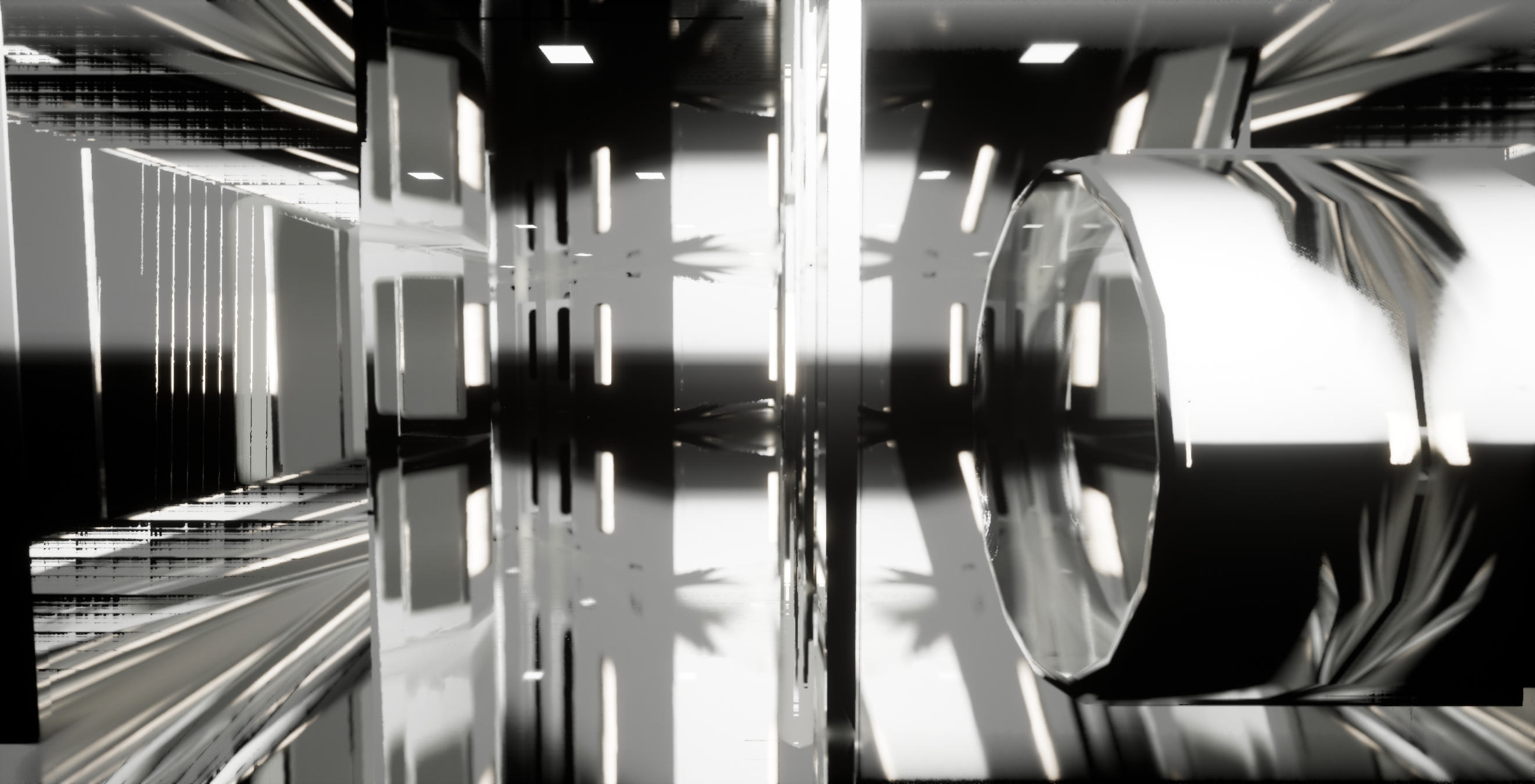}
	}
}
\caption{Realistic rendering of a gauge in a virtual environment (left) and model aspects that are considered when producing the final image (right).}
\label{fig:realisticrendering}
\end{figure}

An example of a game engine which is especially capable of realistic rendering compared to similar applications is the Unreal Engine (v. 5.1.1)\cite{unrealengine,2017vsmidComparison}. Its dynamic Lumen lighting system allows the realistic illumination of a synthetic scene including light scattering and reflection starting from different light sources. Many properties of light sources are described using real existing parameters for light (e.g. intensity in lux, light temperature in Kelvin), which simplifies the comparison to real light sources. The lighting model is extended from the material system used in the Unreal Engine. This includes the description of additional properties of surfaces at a certain point besides the pure coloring, such as the refraction of light and the metal and specular reflection. Figure \ref{fig:realisticrendering} demonstrates the usage of the lighting and material system.

To model haze/smoke within the Unreal Engine, two components of the Unreal Engine are used. By utilizing the \textit{Exponential Height Fog} component, haze can be created. It is a homogeneous layer, which limits visibility based on how far objects are away from the scene camera. Since smoke is never that uniform in reality, Unreal Engine's particle system is additionally used to generate smoke clouds between the camera and gauge. The combination of both features allows for the realistic perception of haze- and smoke-filled environments while still being able to control the amount of it freely.

To generate images of different gauges within the environment, a \textit{Cine Camera Actor} is used whose position and parameters (e.g., aspect ratio, lens settings, focus) can be freely adjusted and which mimics functions of a real cameras as well (e.g., auto-exposure). To automate camera movement and export of images, the \textit{EasySynth} Plugin is used. It moves the camera along a predefined path and takes images at specific points. Combining this route with a specific set of parameter values for the smoke system and illumination, the goal of creating images of gauges in quantifiable haze- and smoke-filled environments under different illumination and camera conditions is achieved. A subset of the resulting data set (varying the smoke intensity) can be viewed in Sect.~\ref{sec:datasets}. %Figure \ref{fig:realisticrendering}.

%\begin{figure}
%\centering
%\begin{tabular}{ccc}
%	\includegraphics[width=0.33\linewidth]{./images/figures/img64} & \includegraphics[width=0.33\linewidth]{./images/figures/img65} & \includegraphics[width=0.33\linewidth]{./images/figures/img66} \\
%	\includegraphics[width=0.33\linewidth]{./images/figures/img67} & \includegraphics[width=0.33\linewidth]{./images/figures/img68} & \includegraphics[width=0.33\linewidth]{./images/figures/img69} \\
%	\includegraphics[width=0.33\linewidth]{./images/figures/img70} & \includegraphics[width=0.33\linewidth]{./images/figures/img71} & \includegraphics[width=0.33\linewidth]{./images/figures/img72} \\
%	\includegraphics[width=0.33\linewidth]{./images/figures/img73} & \includegraphics[width=0.33\linewidth]{./images/figures/img74} &  \\
%\end{tabular}
%\caption{Subset of the synthetic dataset showing a view on gauges. The smoke intensity is linearly increased in the images from top left to bottom right from a clear image to the maximum intensity.}
%\label{fig:realisticrendering}
%\end{figure}
%%%%%%%%%%%%
%%%%%%%%%%%%

\subsubsection{Gauge datasets}\label{sec:datasets}

\begin{figure}[h!]
 \centering
 \includegraphics[scale=0.40]{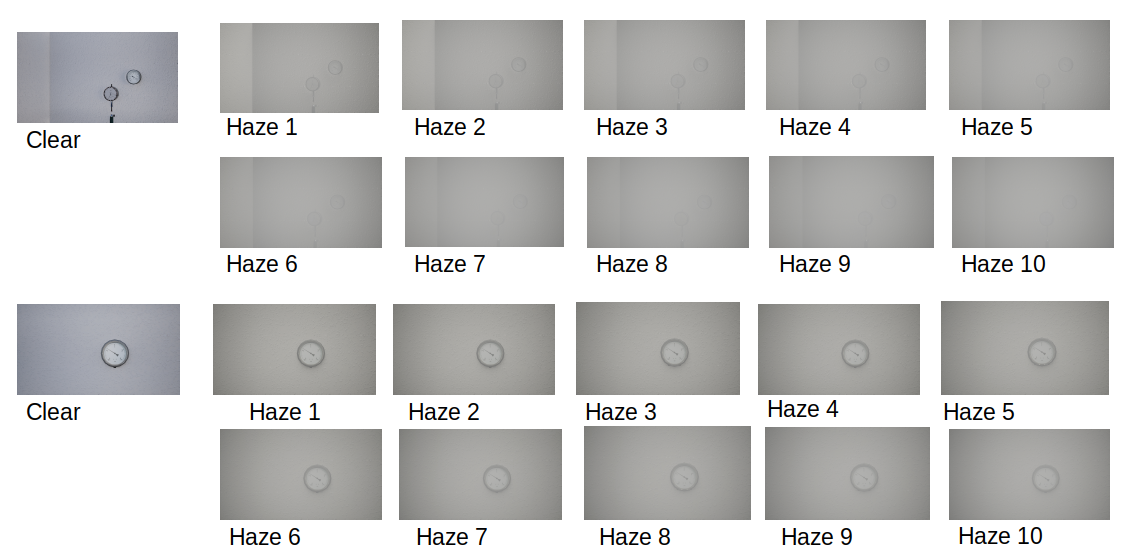}
 \caption{Two examples from the \textbf{Synthetic Haze} dataset.}
 \label{fig:synthetic_haze_ds}
 \end{figure}

\begin{figure}[h!]
 \centering
 \includegraphics[scale=0.35]{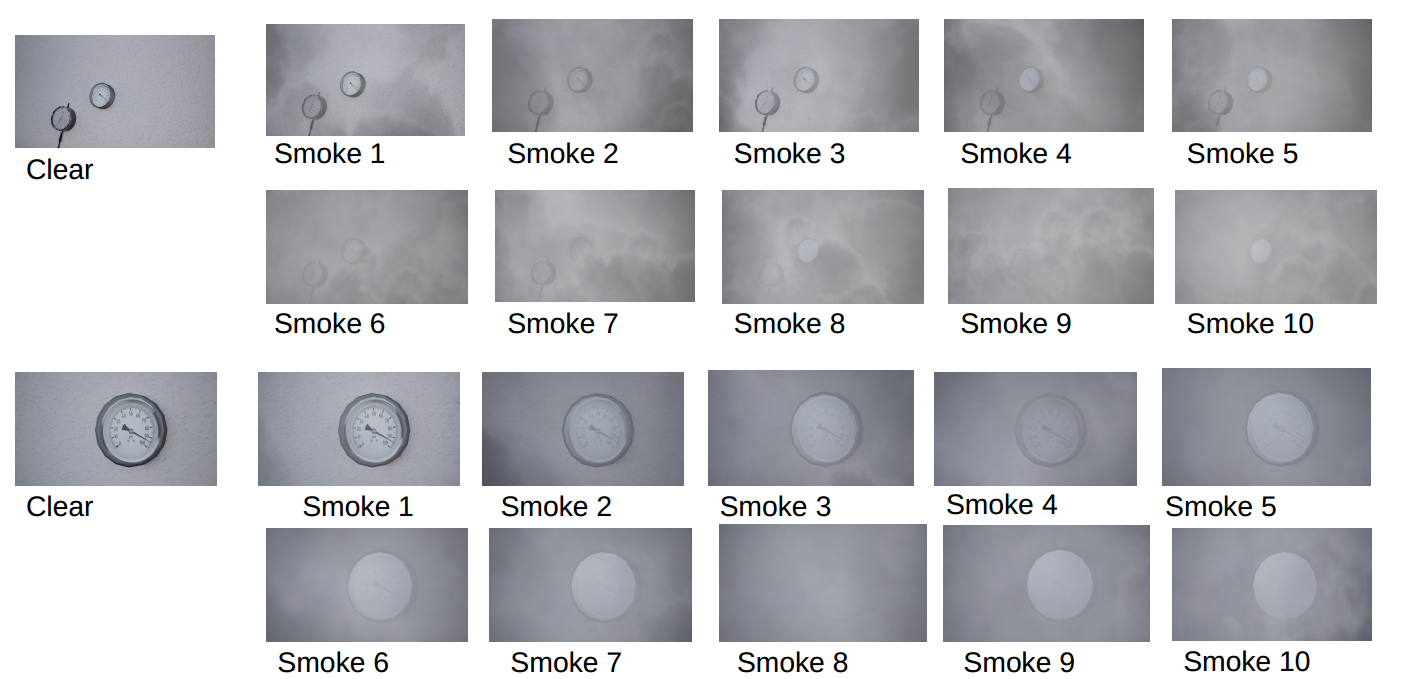}
 \caption{Two examples from the \textbf{Synthetic Smoke} dataset.}
 \label{fig:synthetic_smoke_ds}
 \end{figure}

By means of the Unreal Engine, it has been created two datasets; i.e. a hazy and a smoky. Figures~\ref{fig:synthetic_haze_ds} and~\ref{fig:synthetic_smoke_ds} show two different examples of the synthetic haze and smoke cases. The ground truth (i.e. clear) is presented with the ten different levels of haze and smoke, respectively.\newline

%------------------------------------------------------------------------------
 %-------------------------------------
 %-------------------------------------

For training purposes, the train, validation, and test split is used as shown in Table\ref{tab:synthetic_haze} and~\ref{tab:table_synthetic_smoke}. Two separate datasets, namely haze dataset and smoke dataset are used to train two models each for the FFA-Net and AECR-Net.\newline
The training (train), validation (val), and testing (test) sets split was considered to be 80/10/10 percent, respectively.

\begin{table}[h!]
\begin{minipage}{0.5\linewidth}
\centering
\begin{tabular}{ |p{3cm}||p{3cm}|  }
\hline
\multicolumn{2}{|c|}{\textbf{Synthetic Haze Dataset}} \\ \hline \hline
\multicolumn{2}{|c|}{Train} \\ \hline
Clear Images & Hazy Images \\ \hline
350 & 3500 \\ \hline
\multicolumn{2}{|c|}{Val} \\ \hline
Clear Images & Hazy Images\\ \hline
43 & 430 \\ \hline
\multicolumn{2}{|c|}{Test} \\ \hline
Clear Images & Hazy Images\\ \hline
43 & 430 \\ \hline
\multicolumn{2}{|c|}{Total: 4796 images} \\ \hline
\end{tabular}
\caption{Number of images on the training, validation, and testing sets of the synthetic Haze dataset.}
\label{tab:synthetic_haze}
\end{minipage}%
\begin{minipage}{0.5\linewidth}
\centering
\begin{tabular}{ |p{3cm}||p{3cm}|  }
\hline
\multicolumn{2}{|c|}{\textbf{Synthetic Smoke Dataset}} \\ \hline \hline
\multicolumn{2}{|c|}{Train} \\ \hline
Clear Images & Smoky Images\\ \hline
769 & 7690 \\ \hline
\multicolumn{2}{|c|}{Val} \\ \hline
Clear Images & Smoky Images\\ \hline
95 & 950 \\ \hline
\multicolumn{2}{|c|}{Test} \\ \hline
Clear Images & Smoky Images\\ \hline
95 & 950 \\ \hline
\multicolumn{2}{|c|}{Total: 9590 images} \\ \hline
\end{tabular}
\caption{Number of images on the training, validation, and testing sets of the synthetic Smoke dataset.}
\label{tab:table_synthetic_smoke}
\end{minipage}
\end{table}

\subsection{Implementation of methods}\label{sec:implementation_of_methods}
In this paper has been used three methods, namely: BCCR, FFA-Net, and AECR-Net (see Sect.~\ref{sec:sota}) to enhance the vision of hazy/smoky images. The aforementioned works have been selected given their good performance as referred in the literature compared to other networks (see also discussion in section~\ref{sec:discussion}). Although the focus of this work are the deep learning methods; the optimization approach BCCR has also been used in order to compare and to gain insight of the feasibility and reliability of deep learning vs. non-deep learning methods.\newline

\subsubsection{Implementation of the classical method BCCR}\label{sec:implementation_bccr}
The libraries used for obtaining the enhanced image through this method has been obtained from the $pypi$ project page \cite{meng2013bccr_repo}. This efficient image dehazer proposed  by Meng et al. \cite{meng2013bccr} does not make use of any benchmark dataset to train. It enhances the image by means of the regularization method. The ground truth image is used later on to evaluate the output with respect to the PSNR and SSIM metrics. Besides, the default values of the parameters are also adopted from the paper.\newline

\subsubsection{Implementation of the deep learning networks}\label{sec:implementation_dl}

\subsubsection*{Implementation of FFA-Net}\label{sec:implementation_ffanet}

The FFA-Net repository is available in the following GitHub \href{https://github.com/zhilin007/FFA-Net}{repository}. Except for the hyper-parameters shown in Table~\ref{tab:parameters_dl} and that the parameter \textit{residual blocks} adopted here is nineteen, and not twenty (see \cite{qin2019ffanet}), the rest of the hyper-parameters adopted here  are the same as the ones used in the FFA-Net paper. 

\begin{table}[htbp]
\centering
\begin{tabular}{|c|c|}
\hline
\textbf{Parameters} & \textbf{Values} \\    \hline \hline
batch size          & 4      \\    \hline
evaluation step  & 5000  \\    \hline
Epochs             & 100       \\    \hline
%    Image resolution & 1366x768 \\ \hline
\end{tabular}
\caption{Parameters used for training FFA-Net and AECR-Net methods.}
\label{tab:parameters_dl}
\end{table}

\subsubsection*{Implementation of AECR-Net}\label{sec:implementation_aecrnet}
The AECR-Net repository is available in the following GitHub \href{https://github.com/GlassyWu/AECR-Net}{repository}. The original repository interprets the dataset in the \textit{h5} format, here it has been adapted so that it could read other formats such as \textit{png} and \textbf{jpeg}. Except for the hyper-parameters listed in Table~\ref{tab:parameters_dl}, here it has been adopted with the same set-up.\newline

\subsubsection*{Input of the Gauge datasets}\label{sec:input_gauge_dataset}
The synthetic haze and smoke datasets introduced in Sect.~\ref{sec:datasets} are used to feed both networks. As the FFA-Net and AECR-Net were implemented to remove the haze from images, it is expected that the \textit{synthetic Haze} dataset suits better for the task of image quality enhancement. The \textit{synthetic Smoke} dataset constitutes a more realistic  case, though at the same time more difficult for the networks to remove the smoke from the corrupted images. In the section~\ref{sec:results}, the outcome for both datasets for the selected frameworks are presented.\newline

\subsection{Metrics}\label{sec:metrics}
To evaluate the performance of the methods, it will be used the metrics: Peak Signal to Noise Ratio (PSNR) and the Structural Similarity Index (SSIM).
%In the field of dehazing these are the two most used and accepted metrics to evaluate image quality and enhancement in image dehazing.

\subsubsection{Peak Signal to Noise Ratio $-$ PSNR}\label{sec:psnr}
This is a commonly used metric to measure the quality of a dehazed image obtained from a dehazing algorithm, with respect to a ground truth image \cite{PSNR}. PSNR measures the degree of signal distortion between the two images and a higher value of PSNR signifies better quality of the dehazed (or desmoked) image. The PSNR can be calculated as follows:

\begin{equation}
\text{PSNR} = 10\log_{10} \left(\frac{\text{MAX}^2}{\text{MSE}}\right)
\end{equation}

where $\text{MAX}$ is the maximum possible pixel value of the image and $\text{MSE}$ is the mean squared error between the dehazed image ($I_{\text{hazed}}$) and the ground truth ($GT$) image. The MSE is computed as:

\begin{equation}
\text{MSE} = \frac{1}{N}\sum_{i=1}^N (GT_i - I_{\text{hazed},i})^2
\end{equation}

where $N$ is the total number of pixels in the image. The goal is to minimize the MSE value to obtain a high PSNR value.

\subsubsection{ Structural Similarity Index $-$ SSIM}\label{sec:ssim}
When it comes to evaluating the visual quality of dehazed images from the perspective of human perception, the PSNR metric alone is not considered very effective. As a result, researchers have turned to the SSIM \cite{SSIM} as an alternative. This metric compares the ground truth and dehazed images in terms of contrast, luminance, and structure.

The SSIM is calculated by taking into account the means and variances of the restored image and the ground truth image, as well as their cross-variance. Two constants, $c_1$ and $c_2$, are included in the calculation, with default values of 0.01 and 0.03, respectively. The equation for calculating SSIM is given by:

\begin{equation}
\text{SSIM}(GT, Out) = \frac{(2\mu_i\mu_r + c_1)(2\sigma_{ri} + c_2)}{(\mu_i^2 + \mu_r^2 + c_1)(\sigma_i^2 + \sigma_r^2 + c_2)}
\end{equation}

The resulting SSIM score ranges from 0 to 1, with a score of 1 indicating that the two images being compared are identical. Since SSIM is highly sensitive to variations in contrast and illumination, it can effectively identify issues related to dehazing, such as incomplete haze removal or over-saturation of pixels.

\section{Results}\label{sec:results}
This section explores the results for the methods: BCCR, FFA-Net, and AECR-Net by using the obtained synthetic \textit{Haze} and \textit{Smoke} datasets. First, the metrics PSNR and SSIM are presented in Sect.~\ref{results_metric}. Subsequently, some qualitatively results on the testing set for the three methods are illustrated in Sect.~\ref{sec:result_qualitatively}.

\subsection{Metrics: SSIM and PSNR} \label{results_metric}
\subsubsection{Haze}
\begin{figure}[h]
\centering
\includegraphics[scale=0.50]{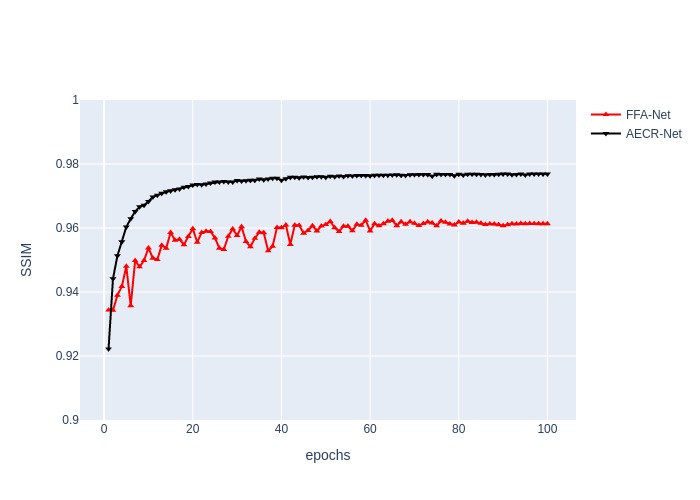}
\caption{SSIM for FFA-Net and AECR-Net trained on \textbf{Synthetic Haze} dataset.}
\label{fig:ssim_haze}
\end{figure}

Figs.~\ref{fig:ssim_haze} shows the results of the similarity index SSIM for the FFA-Net and AECR-Net during training. The maximum SSIM scores are 0.96 and 0.98, respectively. This shows that the latter delivers a slightly better performance. For the BCCR method, the average SSIM score obtained was 0.65. The latter is a non-deep learning method. It is based on an optimization method. It performs poorly when compared to the deep learning methods.\newline

\begin{figure}[h]
\centering
\includegraphics[scale=0.50]{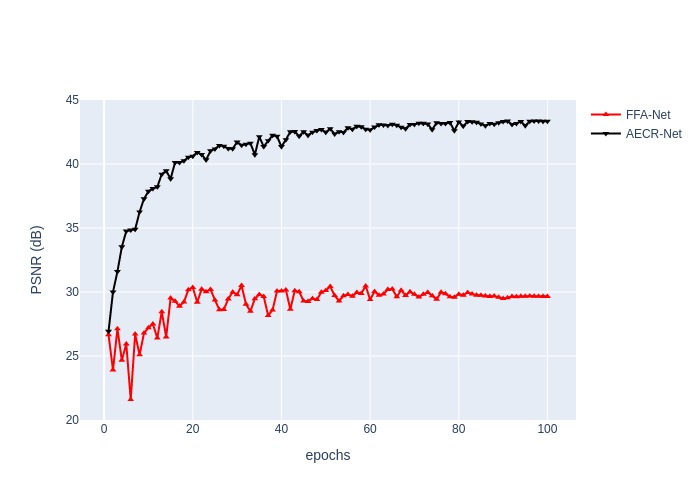}
\caption{PSNR for FFA-Net and AECR-Net trained on \textbf{Synthetic Haze} dataset.}
\label{fig:psnr_haze}
\end{figure}

Fig~\ref{fig:psnr_haze} shows the PSNR result for the FFA-Net and AECR-Net during training. The FFA-Net stabilizes from the epoch forty. The maximum scores achieved are: 30.53\,dB for the FFA-Net and 44\,dB for the AECR-Net. This outcome makes the AECR-Net to stand out in performance with respect to the FFA-Net.  As for the BCCR method, the average PSNR obtained is about 12\,dB. This makes again its performance poorer when compared to non-deep learning methods.\newline

\subsubsection{Smoke}

\begin{figure}[h]
\centering
\includegraphics[scale=0.50]{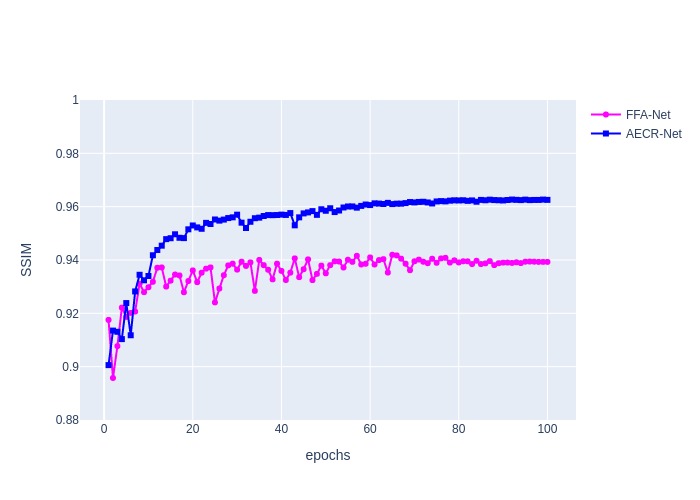}
\caption{SSIM for FFA-Net and AECR-Net trained on \textbf{Synthetic Smoke} dataset.}
\label{fig:ssim_smoke}
\end{figure}

Figure~\ref{fig:ssim_smoke} shows the results of the SSIM score, where the AECR-Net achieved a maximum score of 0.96, which is higher than the FFA-Net's maximum score of 0.94. This indicates that the AECR-Net performs better than the FFA-Net in terms of the image structural similarity. The average SSIM scores for the output images using BCCR method was 0.55.\newline

\begin{figure}[h]
\centering
\includegraphics[scale=0.50]{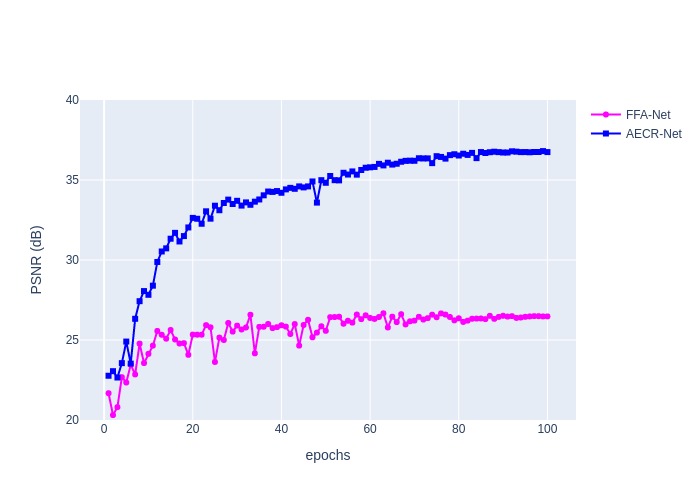}
\caption{PSNR for FFA-Net and AECR-Net trained on \textbf{Synthetic Smoke} dataset.}
\label{fig:psnr_smoke}
\end{figure}

Fig~\ref{fig:psnr_smoke} shows the PSNR outcome. The AECR-Net achieves a maximum PSNR score of 37\,dB, while the FFA-Net's maximum PSNR score is 26\,dB. The average PSNR score for the output images using BCCR method was 9\,dB. These results indicate that the AECR-Net outperforms the FFA-Net in terms of image quality and fidelity, as higher PSNR scores correspond to better image reconstruction accuracy.\newline

\subsection{Qualitatively findings}\label{sec:result_qualitatively}
\subsubsection{Haze Dataset}\label{sec:result_haze}
%The results shown in this section are obtained over synthetic haze dataset. 
%Table~\ref{tab:results_haze_img} summarizes the resulting output images obtained with the three different methods by using three input images with dense haze.\newline

Table~\ref{tab:results_haze_img} displays the results obtained from the three different methods on input images of the testing dataset randomly selected. Every input image-case contains denser levels of haze compared to the previous one. The input images are denoted by a), e), and i), and the corresponding outputs from the BCCR method are denoted by b), f), and j). The BCCR method is observed to produce clear and enhanced outputs; however, they display higher contrast and changed colors compared to the input images, making the output images less natural.\newline

Both the FFA-Net and the AECR-Net methods demonstrate significant haze removal. However, the output images generated by the FFA-Net method, denoted by c), g), and k), exhibit some artifacts. On the other hand, the AECR-Net method produces the best results in this scenario, with color restoration and reduced artifacts observed in the output images, denoted by d), h), and l).

Overall, the results show that, for this particular dataset, the AECR-Net approach removes dense haze from images with the least amount of artifacts and while maintaining color information. The BCCR approach, although delivering outputs that are clear and improved, may not be appropriate if color preservation is significant. \newline
%The FFA-Net approach can be an alternative, although the produced photos might have artifacts.

\begin{table}[h]
\centering
\begin{tabular}{c|ccc}
\textbf{Input} & \multicolumn{3}{c}{\textbf{Results Haze}} \\
\hline
\textbf{Hazy Images} & \textbf{\textit{BCCR}} & \textbf{FFA-Net} & \textbf{AECR-Net} \\
\begin{subtable}{0.22\linewidth}
\centering
\includegraphics[width=\linewidth]{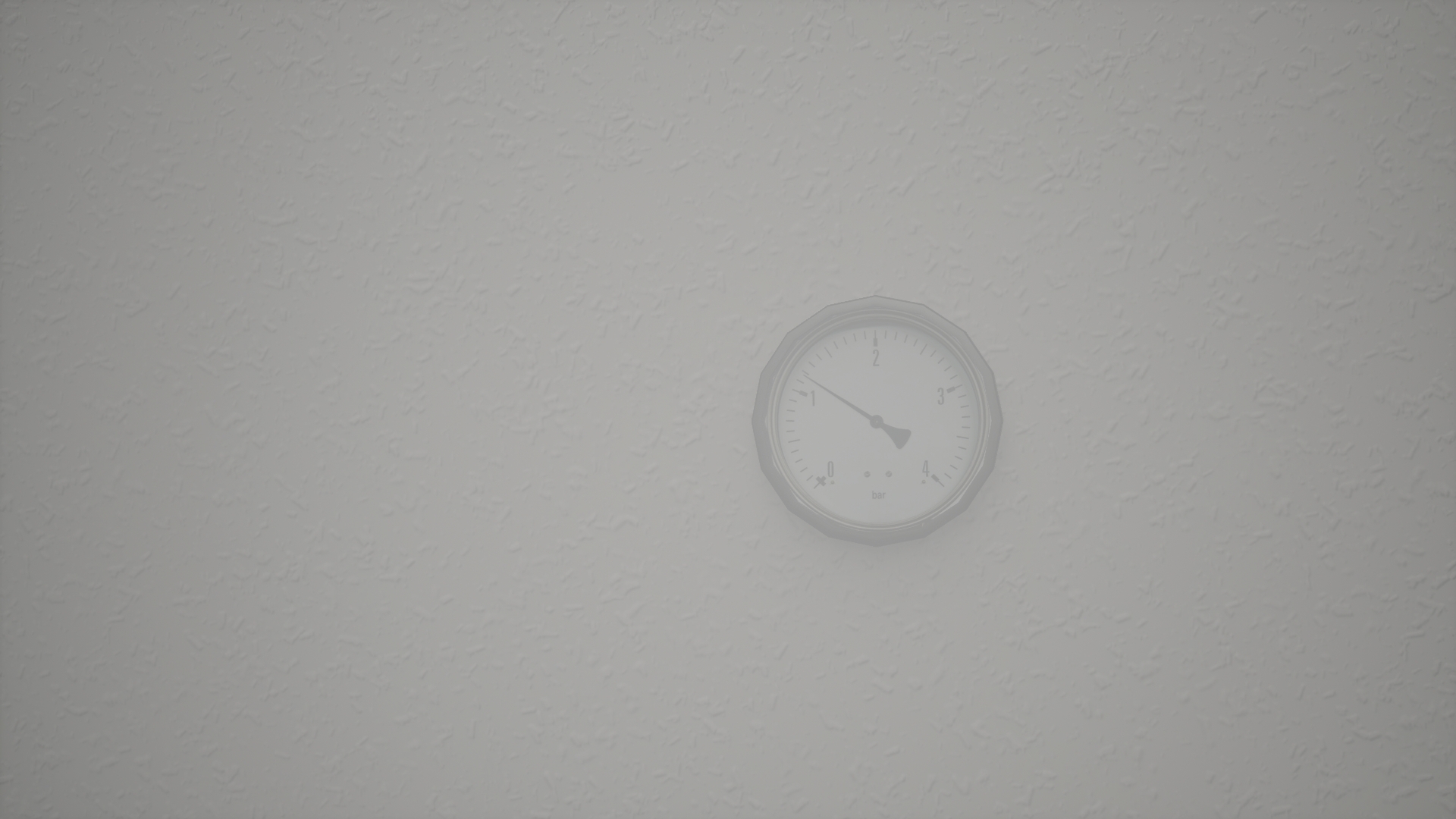} 
\caption{Input 1}
\end{subtable} &
\begin{subtable}{0.22\linewidth}
\centering
\includegraphics[width=\linewidth]{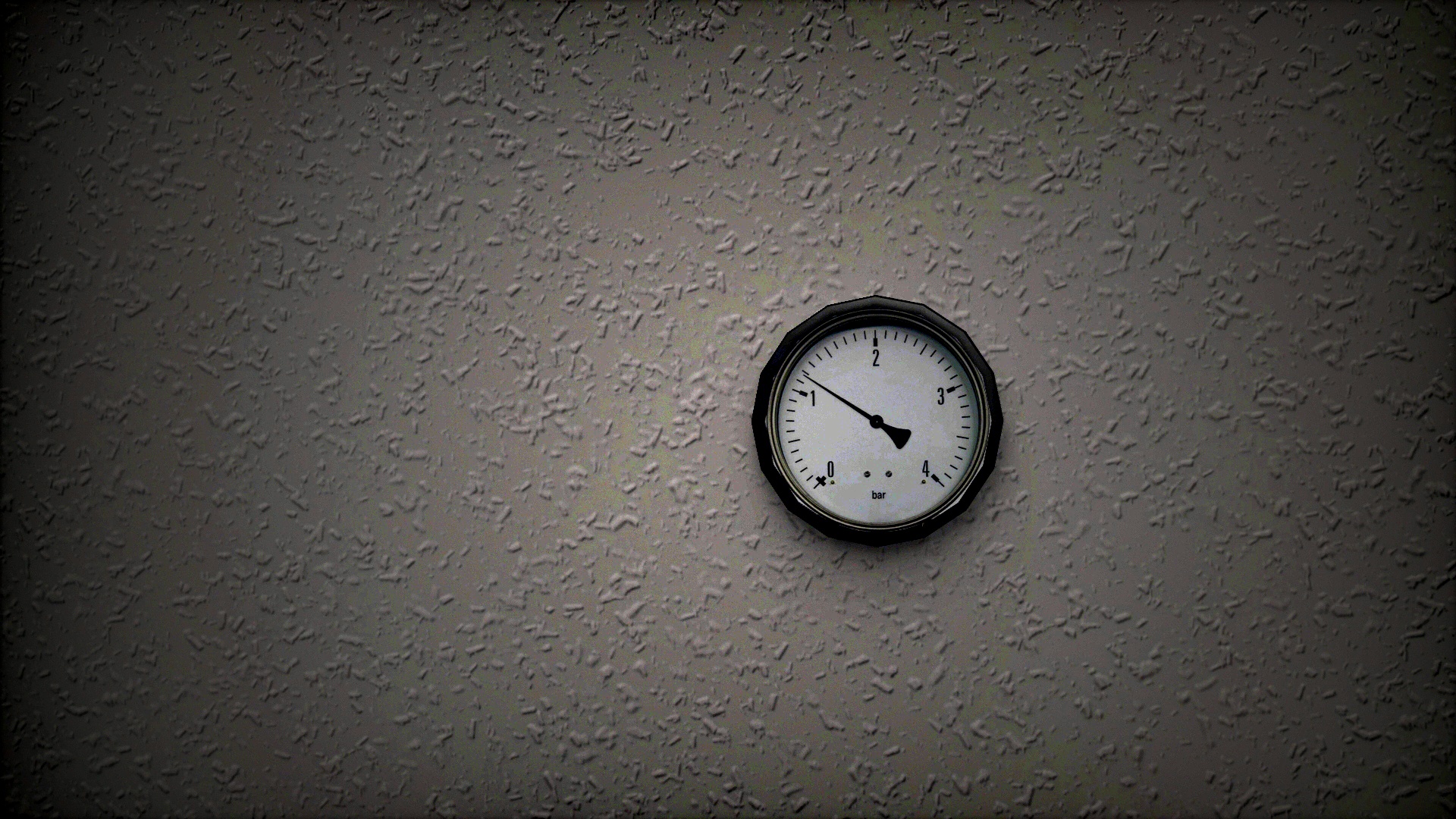}
\caption{BCCR 1}
\end{subtable} &
\begin{subtable}{0.22\linewidth}
\centering
\includegraphics[width=\linewidth]{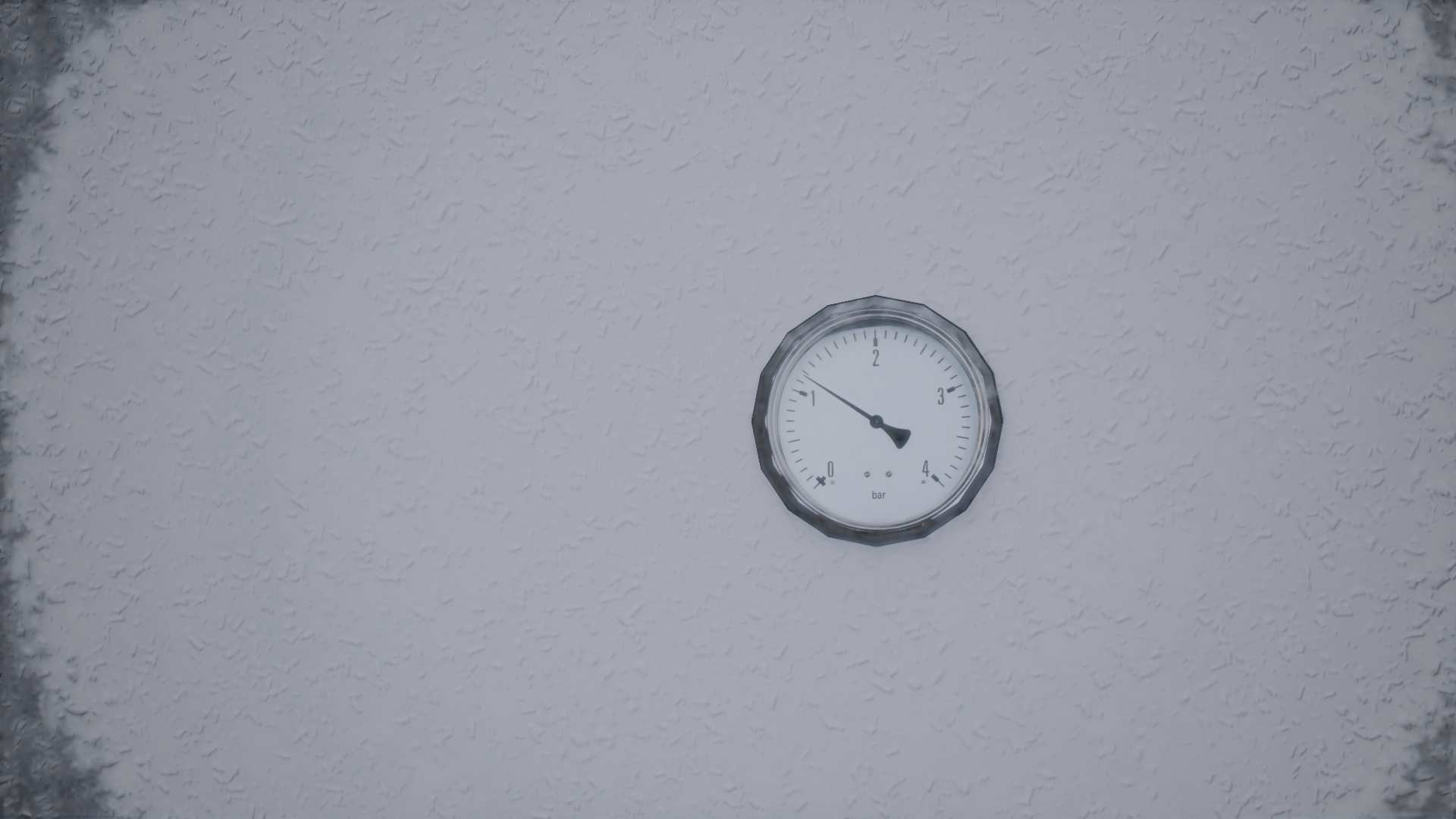}
\caption{FFA-Net 1}
\end{subtable} &
\begin{subtable}{0.22\linewidth}
\centering
\includegraphics[width=\linewidth]{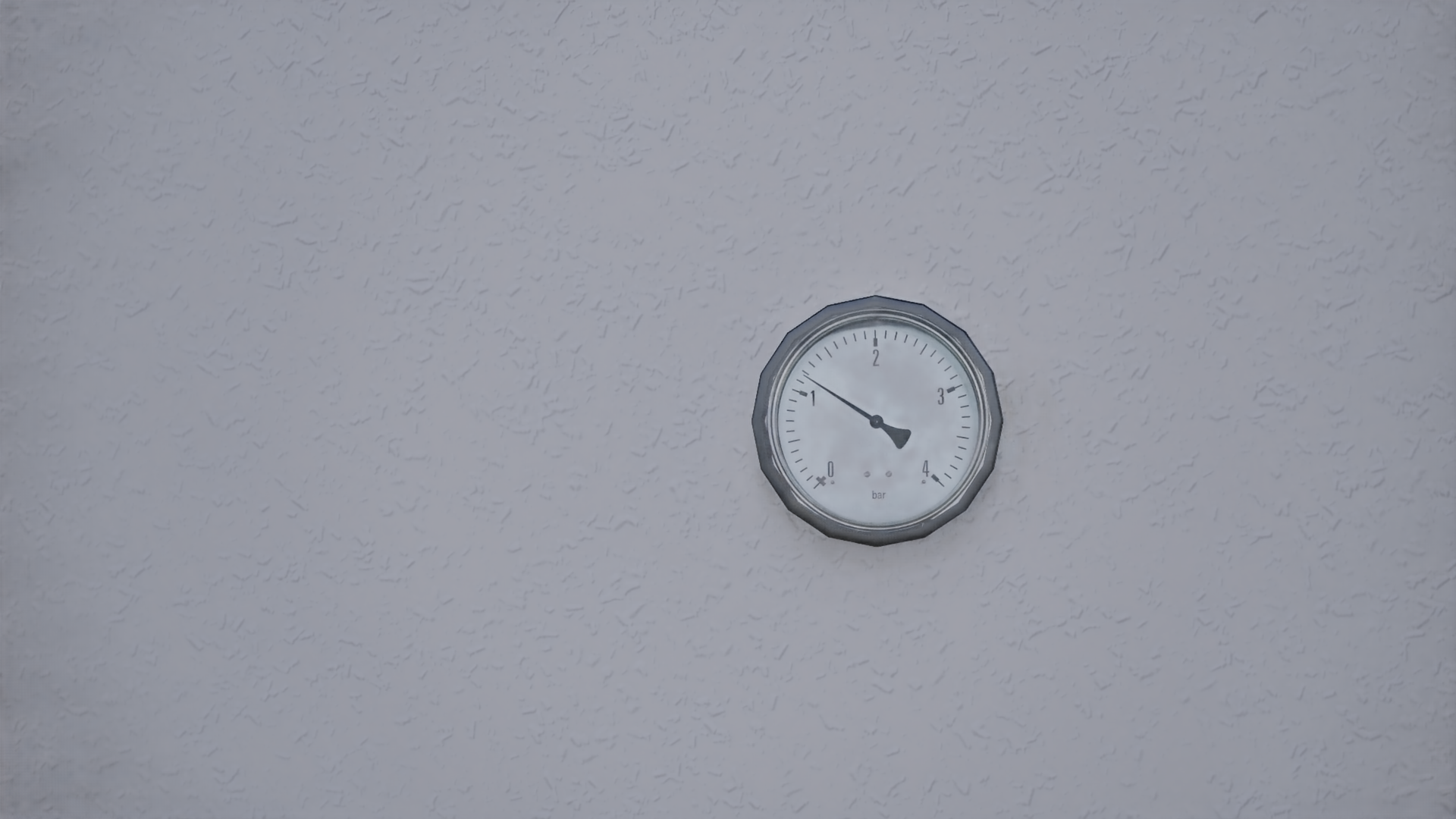}
\caption{AECR-Net 1}
\end{subtable} \\
\begin{subtable}{0.22\linewidth}
\centering
\includegraphics[width=\linewidth]{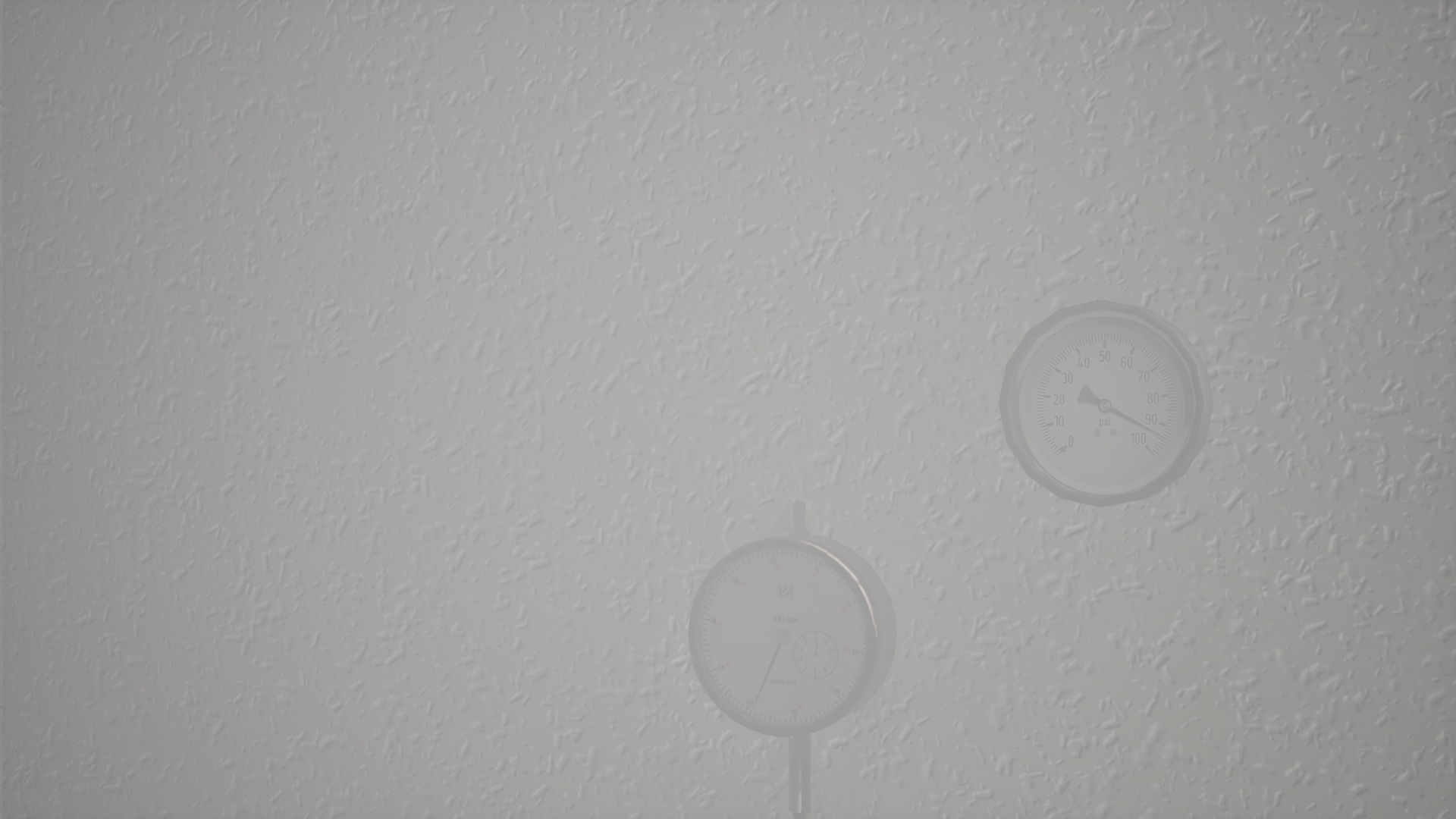}
\caption{Input 2}
\end{subtable} &
\begin{subtable}{0.22\linewidth}
\centering
\includegraphics[width=\linewidth]{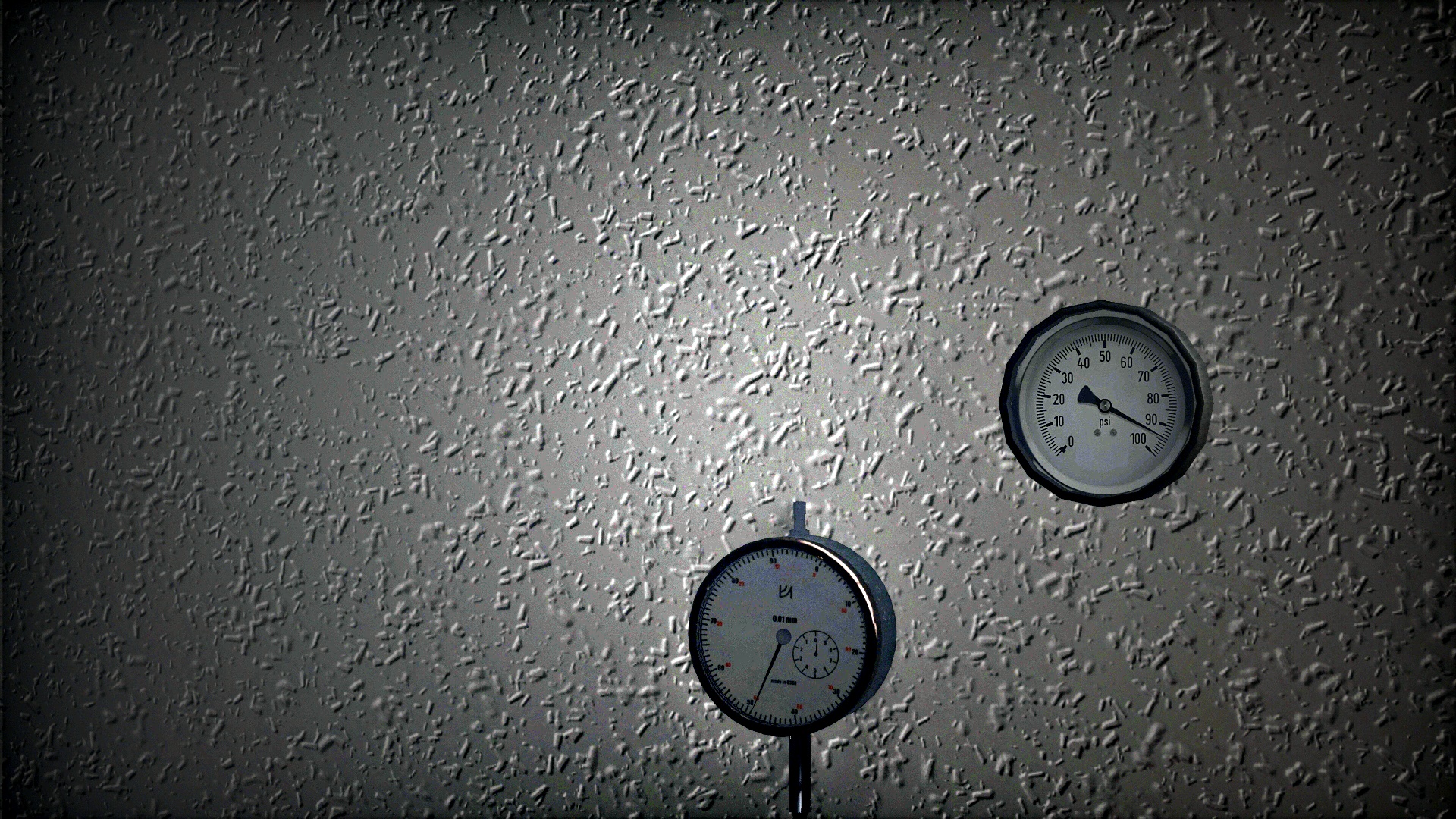}
\caption{BCCR 2}
\end{subtable} &
\begin{subtable}{0.22\linewidth}
\centering
\includegraphics[width=\linewidth]{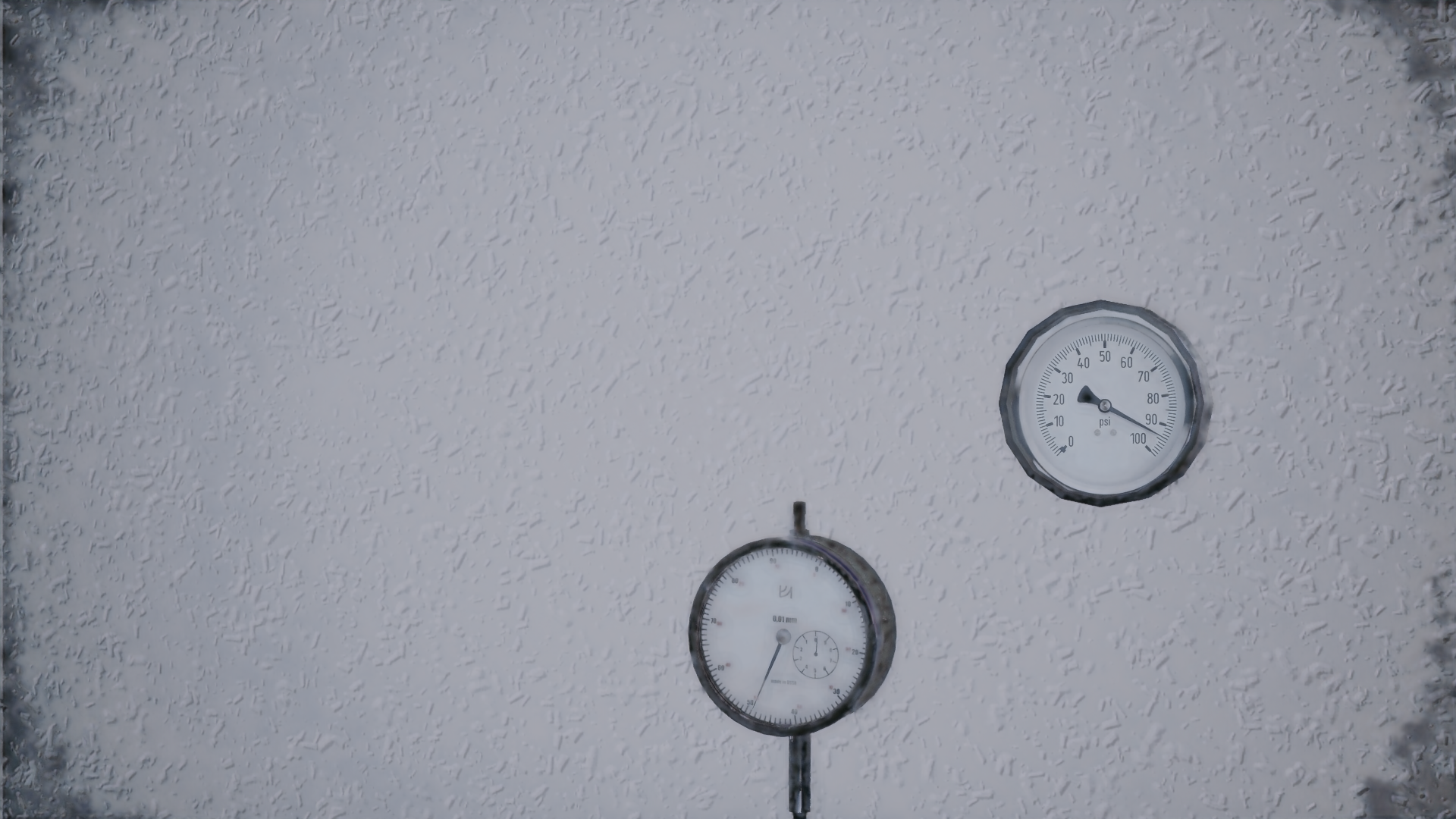}
\caption{FFA-Net 2}
\end{subtable} &
\begin{subtable}{0.22\linewidth}
\centering
\includegraphics[width=\linewidth]{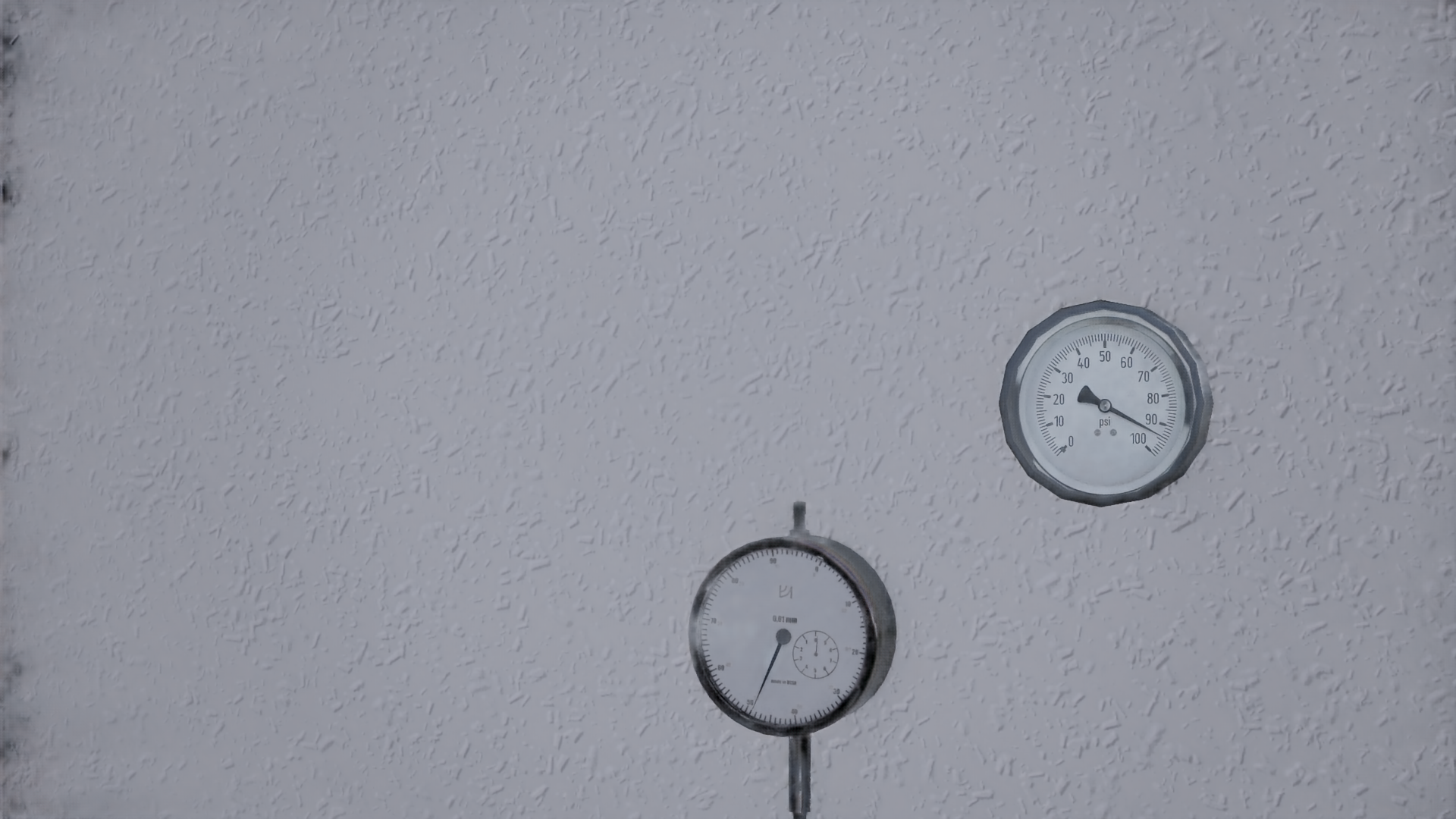}
\caption{AECR-Net 2}
\end{subtable} \\
\begin{subtable}{0.22\linewidth}
\centering
\includegraphics[width=\linewidth]{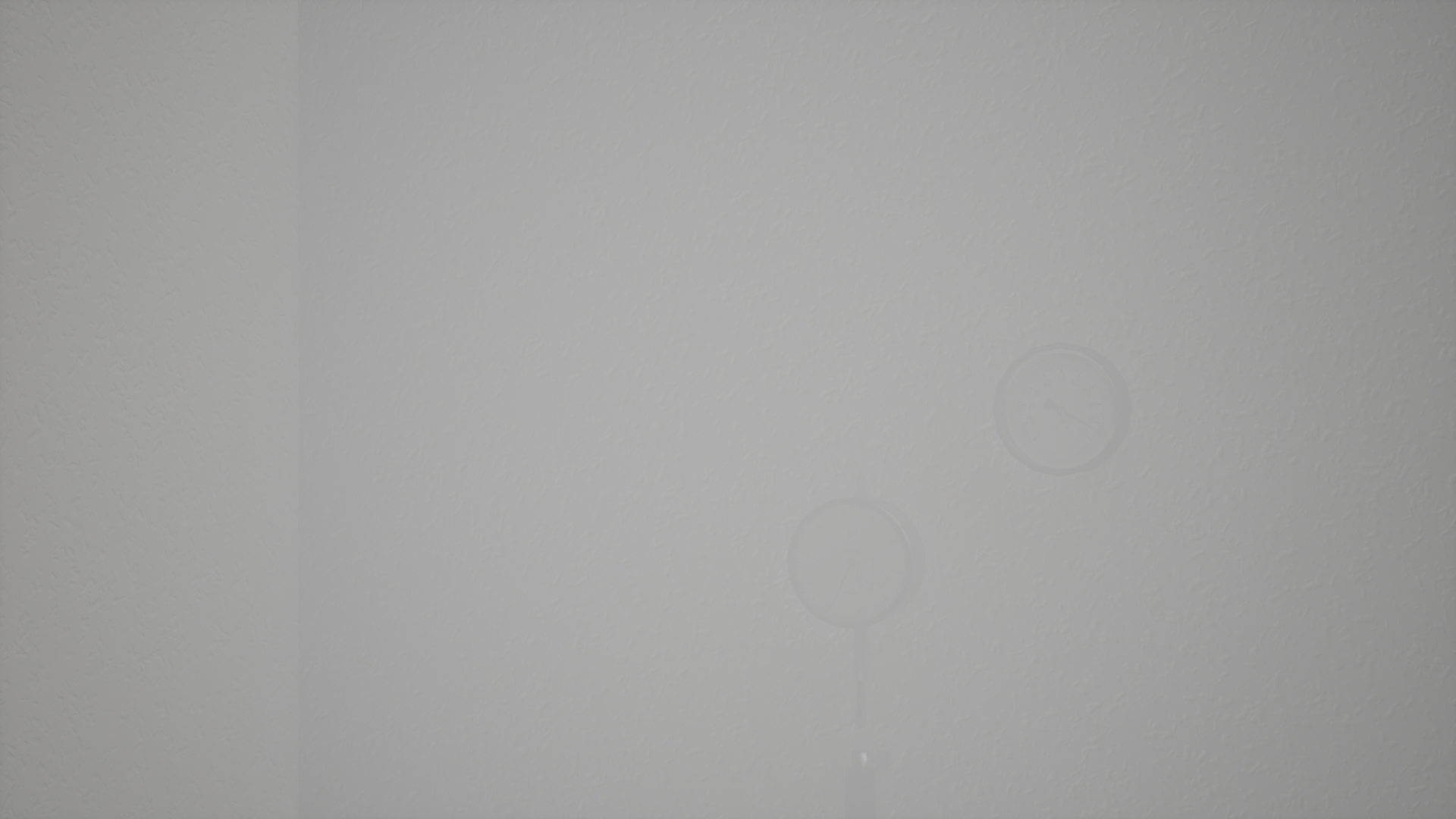}
\caption{Input 3}
\end{subtable} &
\begin{subtable}{0.22\linewidth}
\centering
\includegraphics[width=\linewidth]{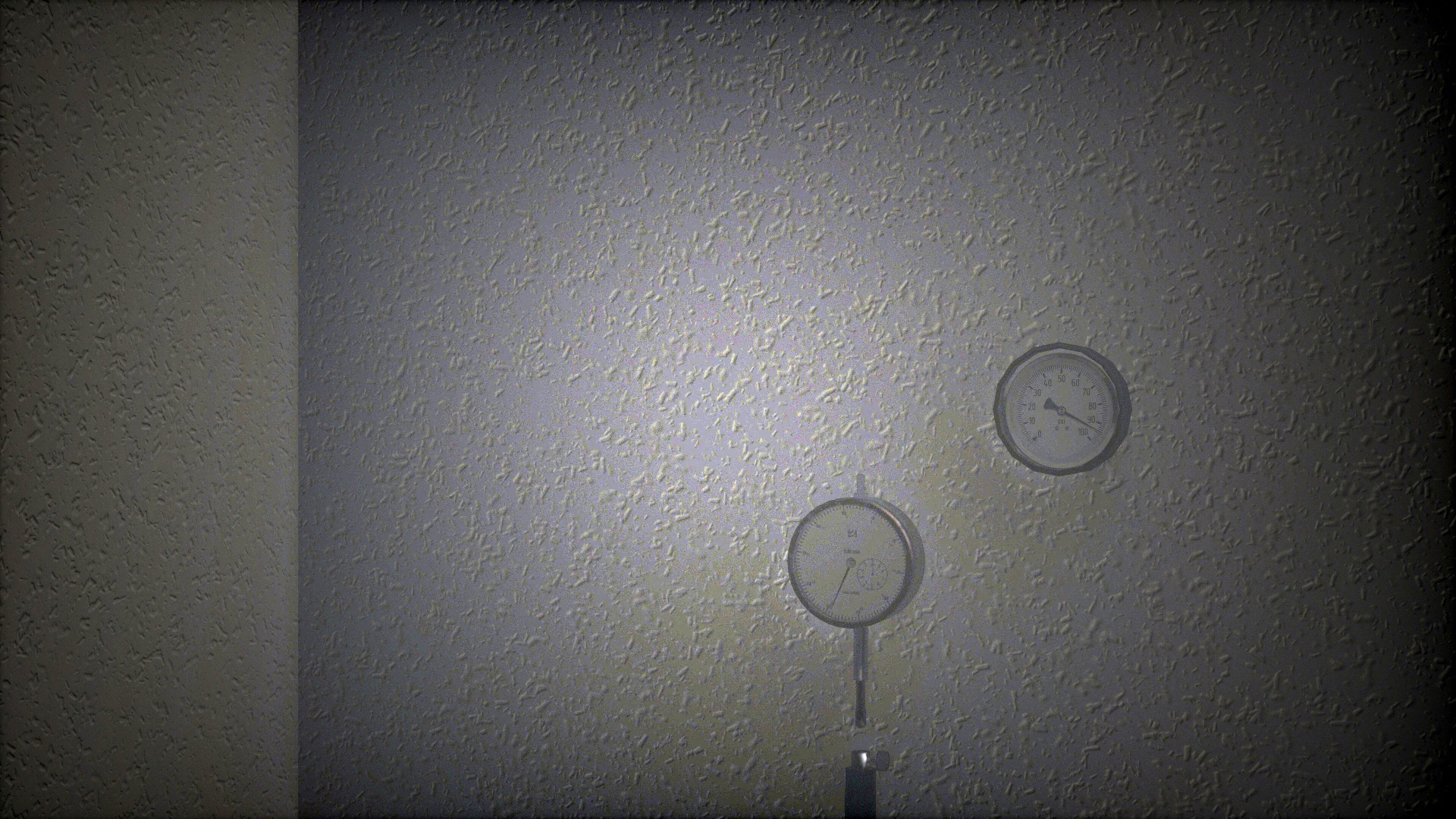}
\caption{BCCR 3}
\end{subtable} &
\begin{subtable}{0.22\linewidth}
\centering
\includegraphics[width=\linewidth]{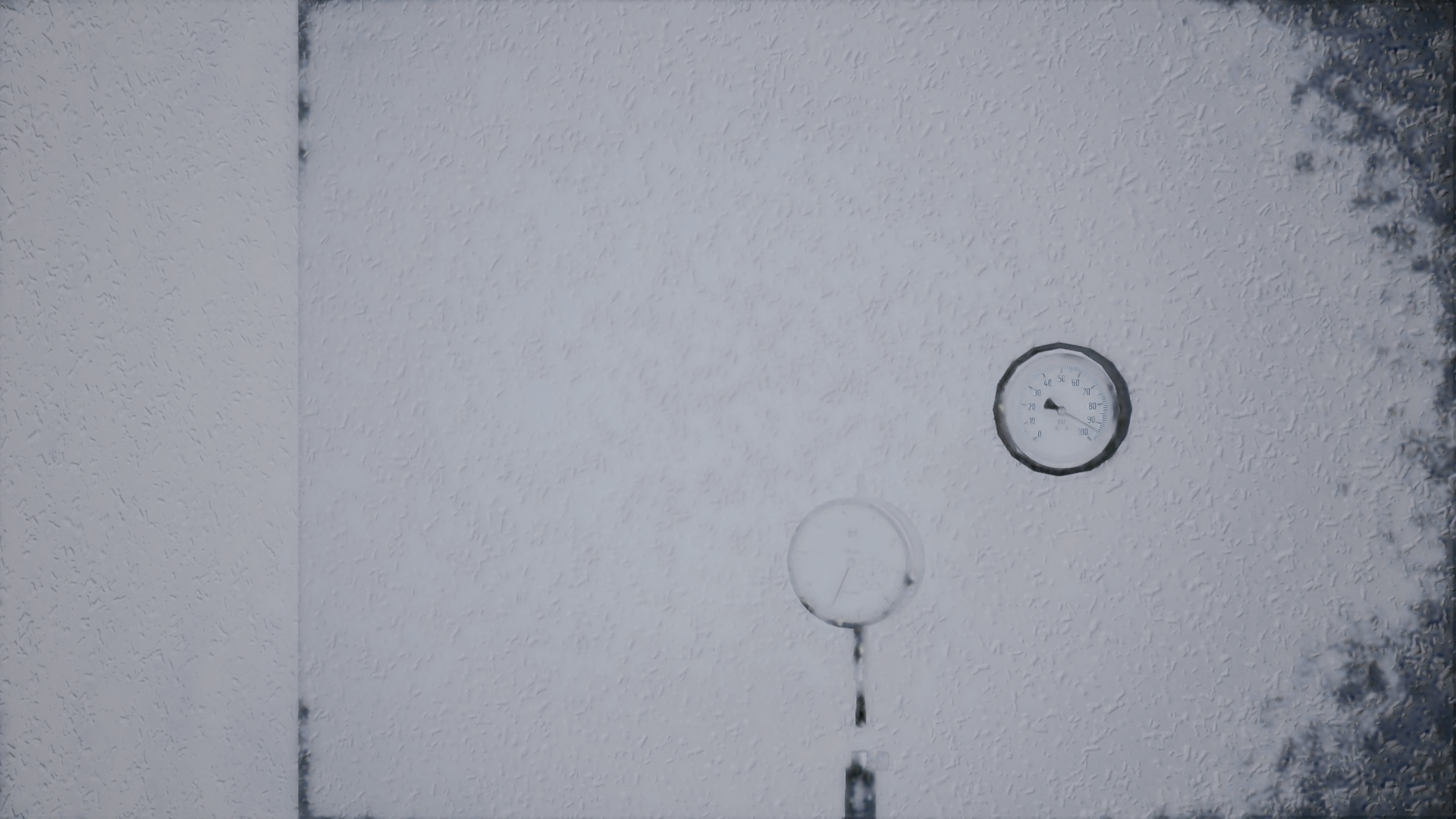}
\caption{FFA-Net 3}
\end{subtable} &
\begin{subtable}{0.22\linewidth}
\centering
\includegraphics[width=\linewidth]{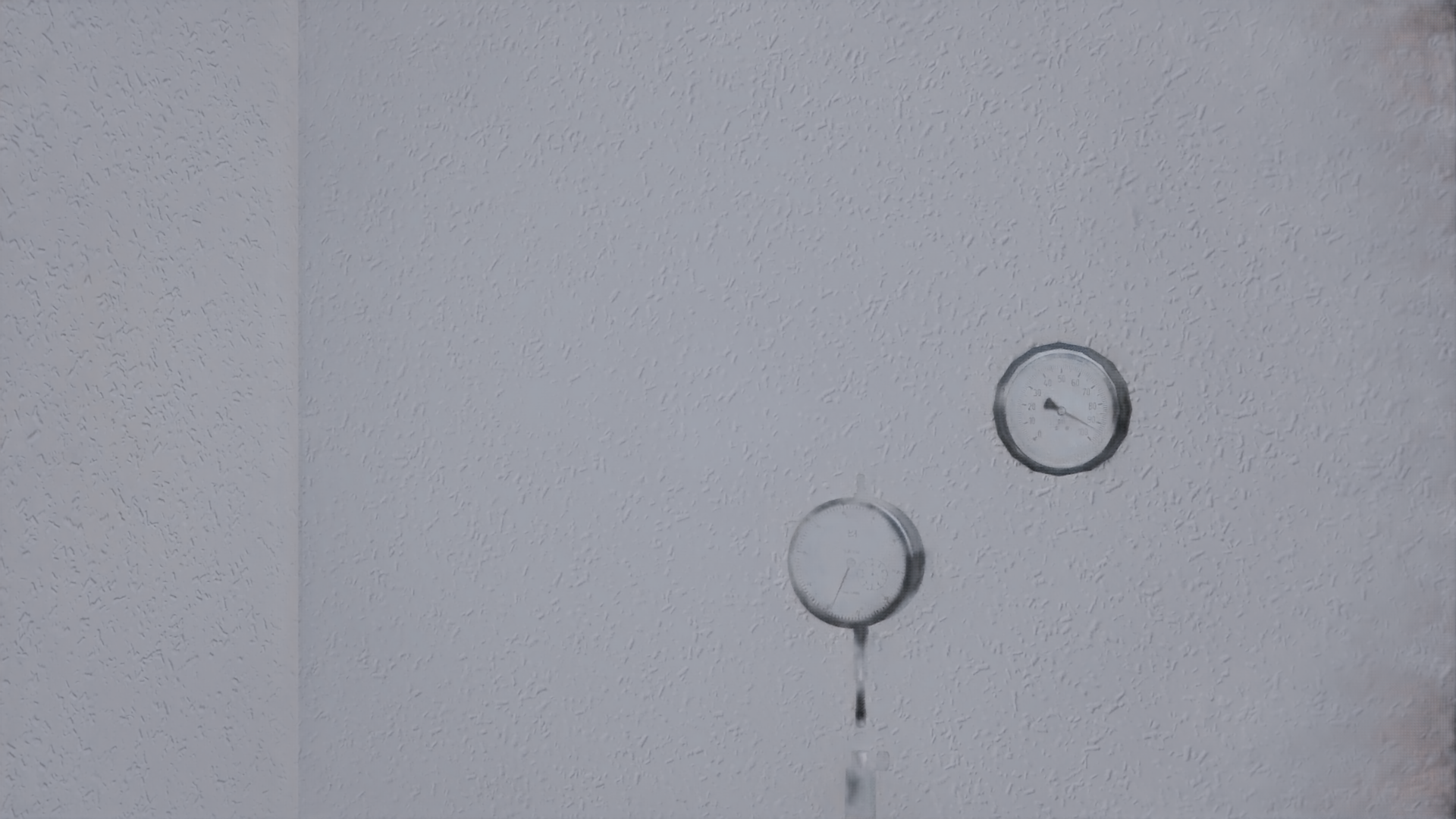}
\caption{AECR-Net 3}
\end{subtable} \\
\end{tabular}
\caption{Results of the three methods tested on three sample hazy images from the \textbf{synthetic haze} dataset. \textbf{First Column:} three input images with dense smoke, \textbf{second column:} output images from BCCR method, \textbf{third column:} output images from FFA-Net, \textbf{fourth column:} output images from AECR-Net.}
\label{tab:results_haze_img}
\end{table}
%%%%%%%%%%%%%%%%%%%%%%%%%%%%%%%%%%%%%%%%%%%%%%%%%%%%%%%%%%%%%%%%%%%%%%%%%%%%%%%%%%%%%%%%%%%%%%

\subsubsection{Smoke Dataset}\label{sec:result_smoke}
In this section, we present the results obtained from experiments conducted on a synthetic Smoke dataset. The objective was to evaluate the performance of three different methods, namely BCCR, FFA-Net, and AECR-Net, in removing dense smoke from images.\newline

Table~\ref{tab:results_smoke_img} summarizes the results obtained by applying the three methods to three different input images from the testing dataset randomly selected. The input images have different varying levels of dense smoke. The input images denoted by a), e), and i) contain heavy smoke, with i) being the densest where the gauge is not visible to the naked eye.\newline

The output images produced by the BCCR method b), f), and j) were not clear and the dense smoke was not completely removed. Moreover, the output images from this method had color discoloration, making them unsuitable for practical post-processing applications. In fact, for the densest smoke, the output image j) amplified the dense smoke from the input image, making the results even worse. On the other hand, the deep learning-based methods, FFA-Net and AECR-Net, produced much clearer and enhanced output images compared to the prior-based method BCCR. However, the output images from the FFA-Net g) and k) were missing some input pixels and still contained some residue of smoke, especially in cases where the input image had very dense smoke.\newline 

The AECR-Net, however, again produced the best results when compared to the BCCR method and the FFA-Net. Output image l) was able to remove the densest smoke from input image i). Although, the image suffers from distortion and artifacts, the output images produced by the AECR-Net try to preserve the colors and contrast levels and minimized artifacts in the output images.\newline

\begin{table}[h]
\centering
\begin{tabular}{c|ccc}
\textbf{Input} & \multicolumn{3}{c}{\textbf{Results Smoke}} \\
\hline
\textbf{Smoky Images} & \textbf{BCCR} & \textbf{FFA-Net} & \textbf{AECR-Net} \\
\begin{subtable}{0.22\linewidth}
\centering
\includegraphics[width=\linewidth]{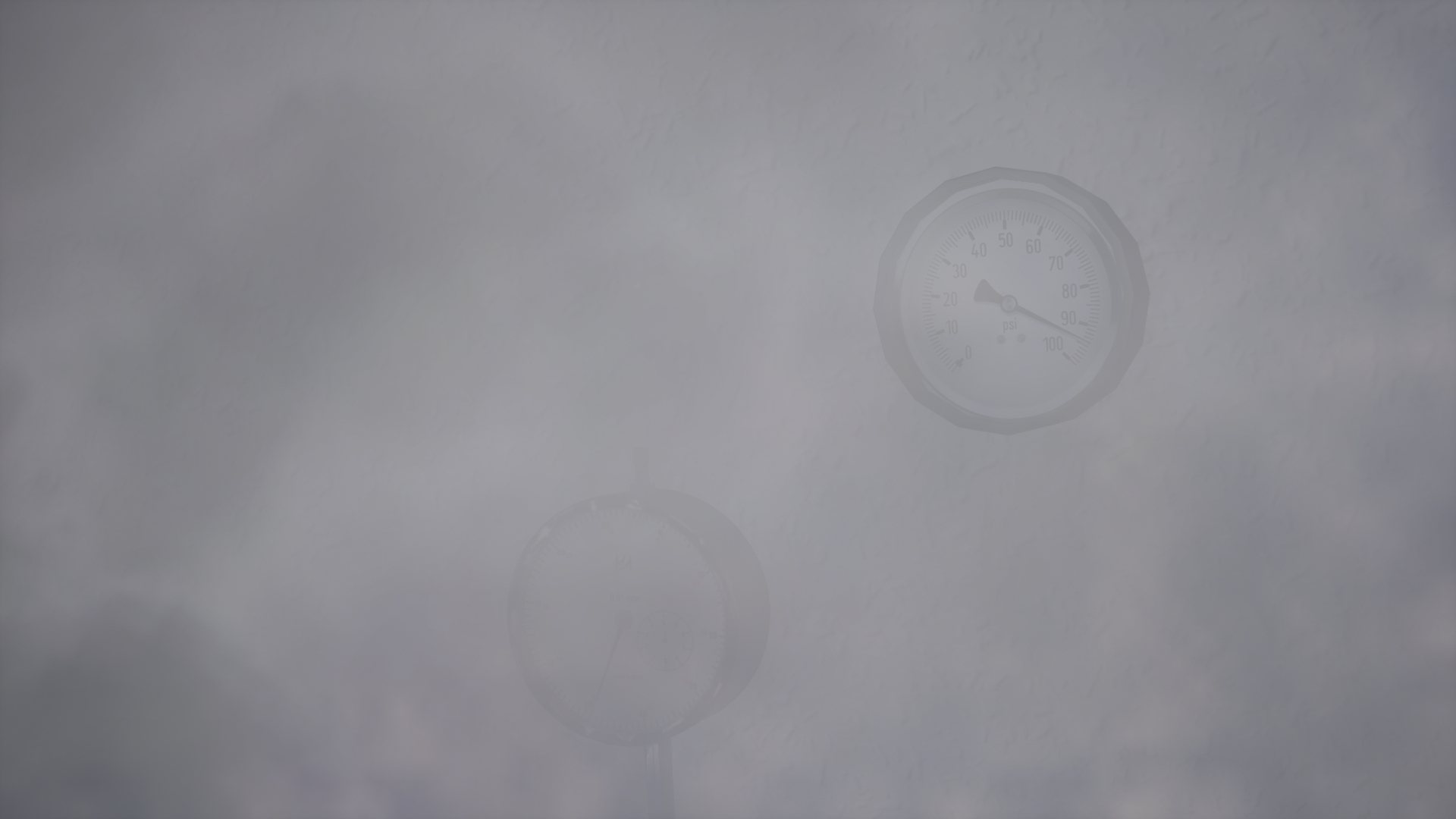}
\caption{Input 1}
\end{subtable} &
\begin{subtable}{0.22\linewidth}
\centering
\includegraphics[width=\linewidth]{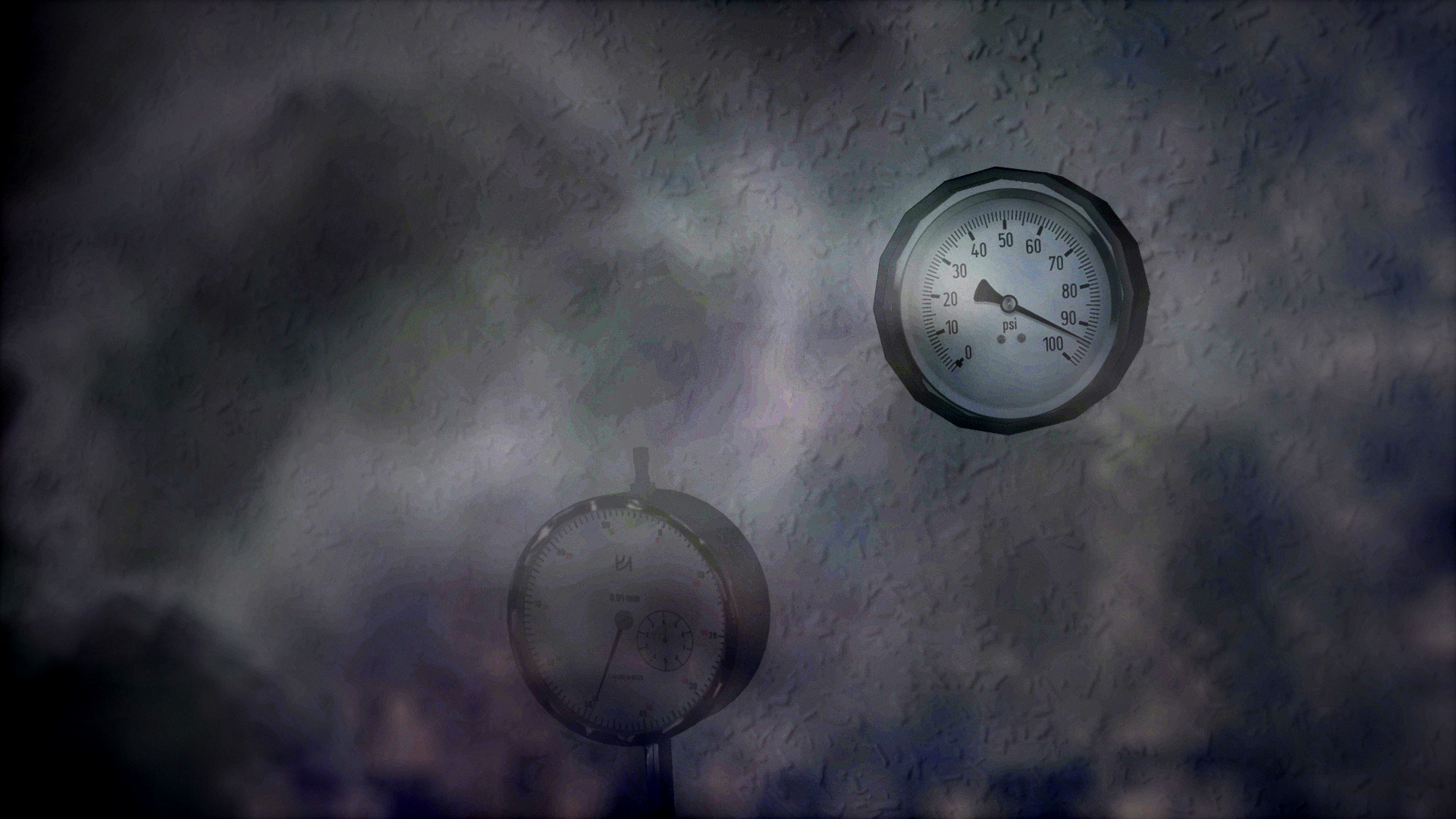}
\caption{BCCR 1}
\end{subtable} &
\begin{subtable}{0.22\linewidth}
\centering
\includegraphics[width=\linewidth]{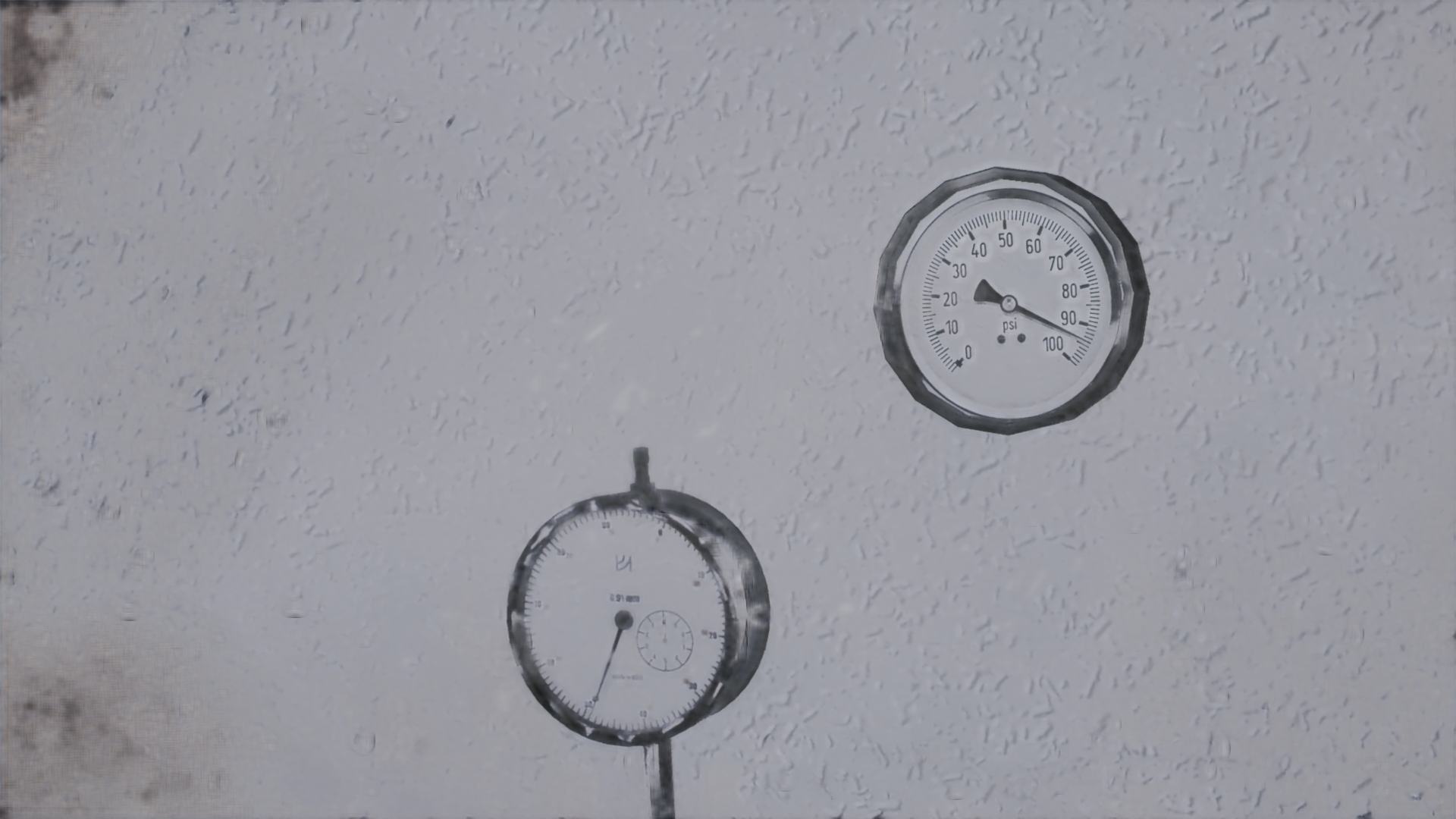}
\caption{FFA-Net 1}
\end{subtable} &
\begin{subtable}{0.22\linewidth}
\centering
\includegraphics[width=\linewidth]{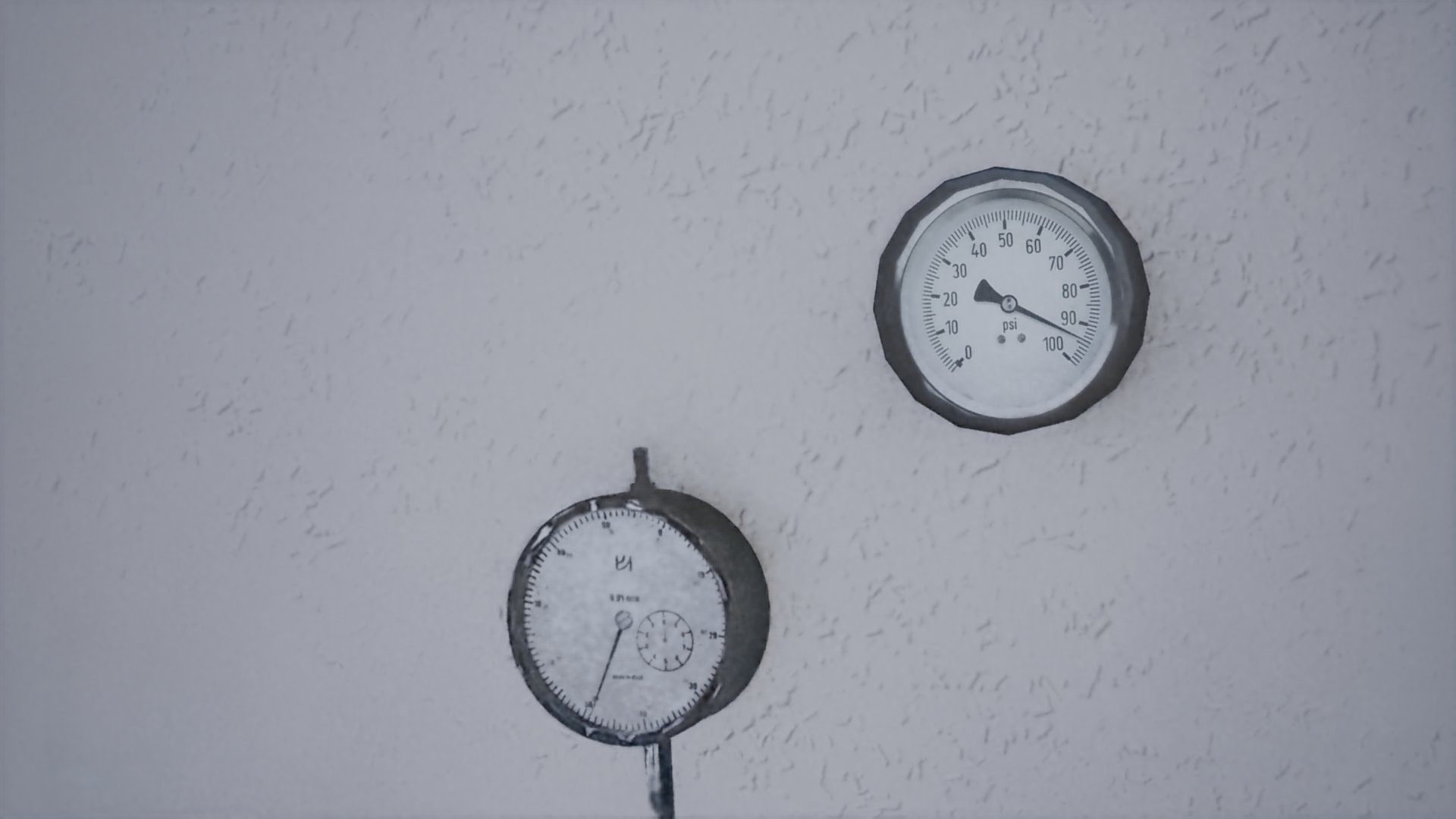}
\caption{AECR-Net 1}
\end{subtable} \\
\begin{subtable}{0.22\linewidth}
\centering
\includegraphics[width=\linewidth]{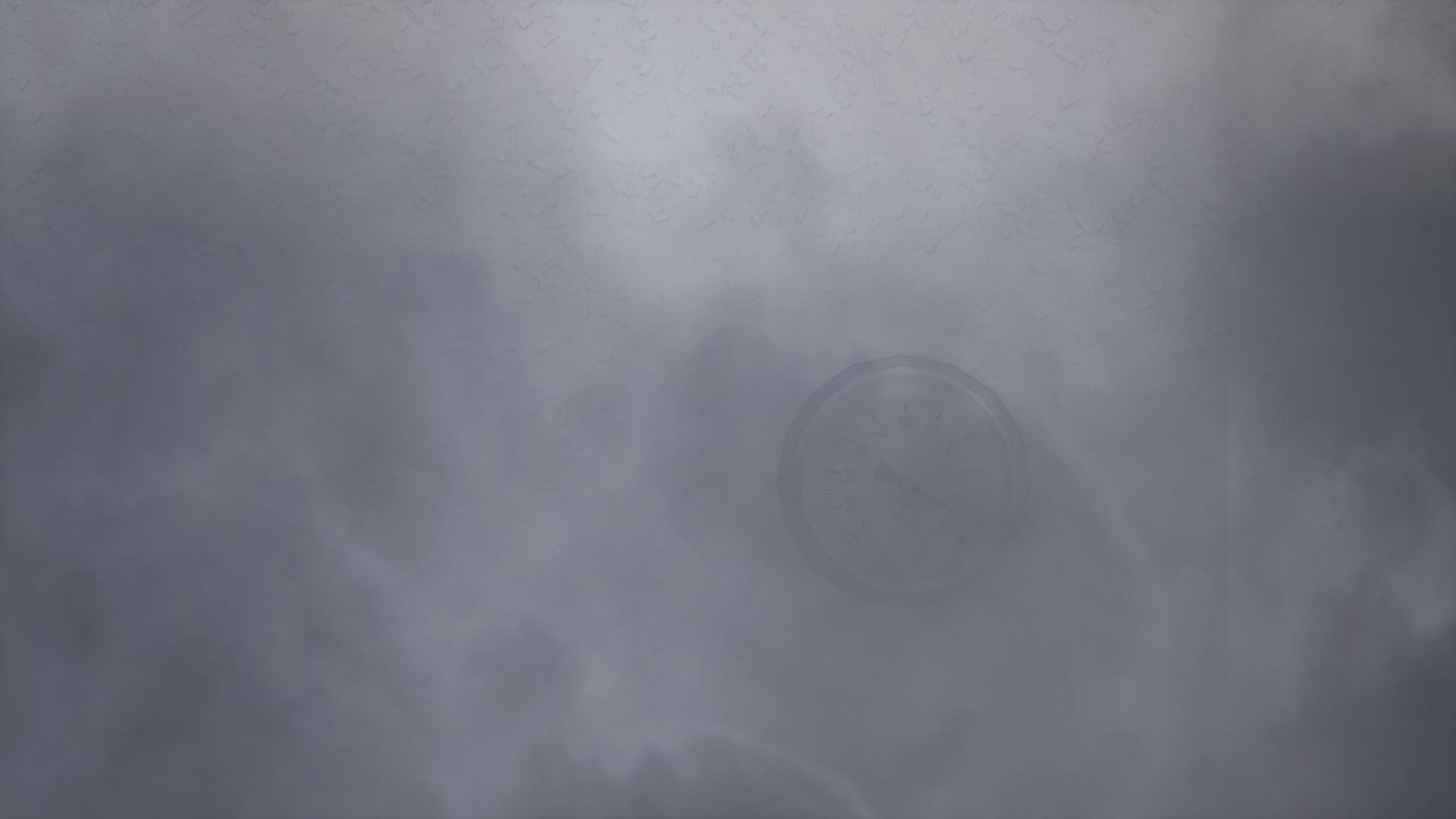}
\caption{Input 2}
\end{subtable} &
\begin{subtable}{0.22\linewidth}
\centering
\includegraphics[width=\linewidth]{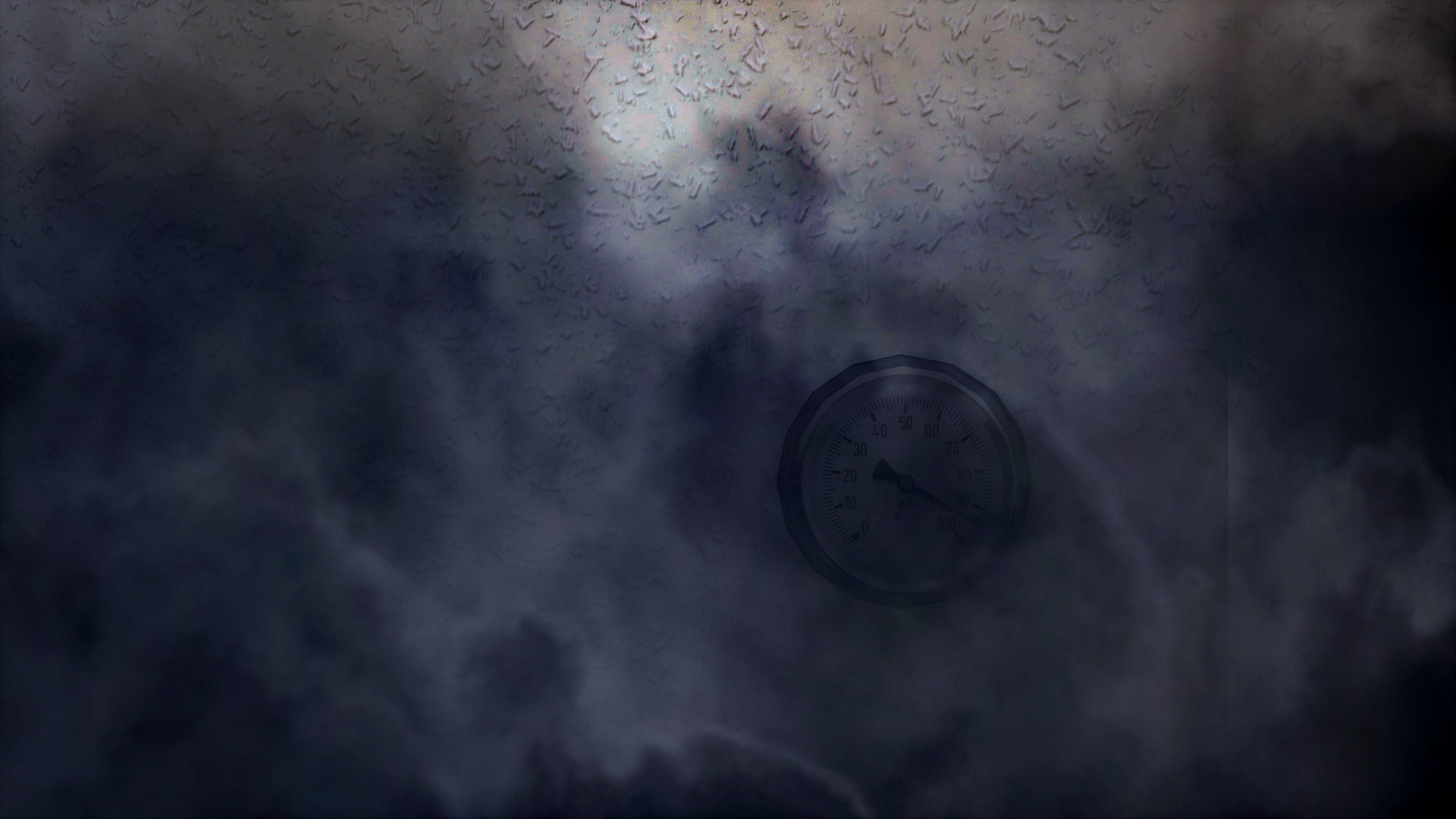}
\caption{BCCR 2}
\end{subtable} &
\begin{subtable}{0.22\linewidth}
\centering
\includegraphics[width=\linewidth]{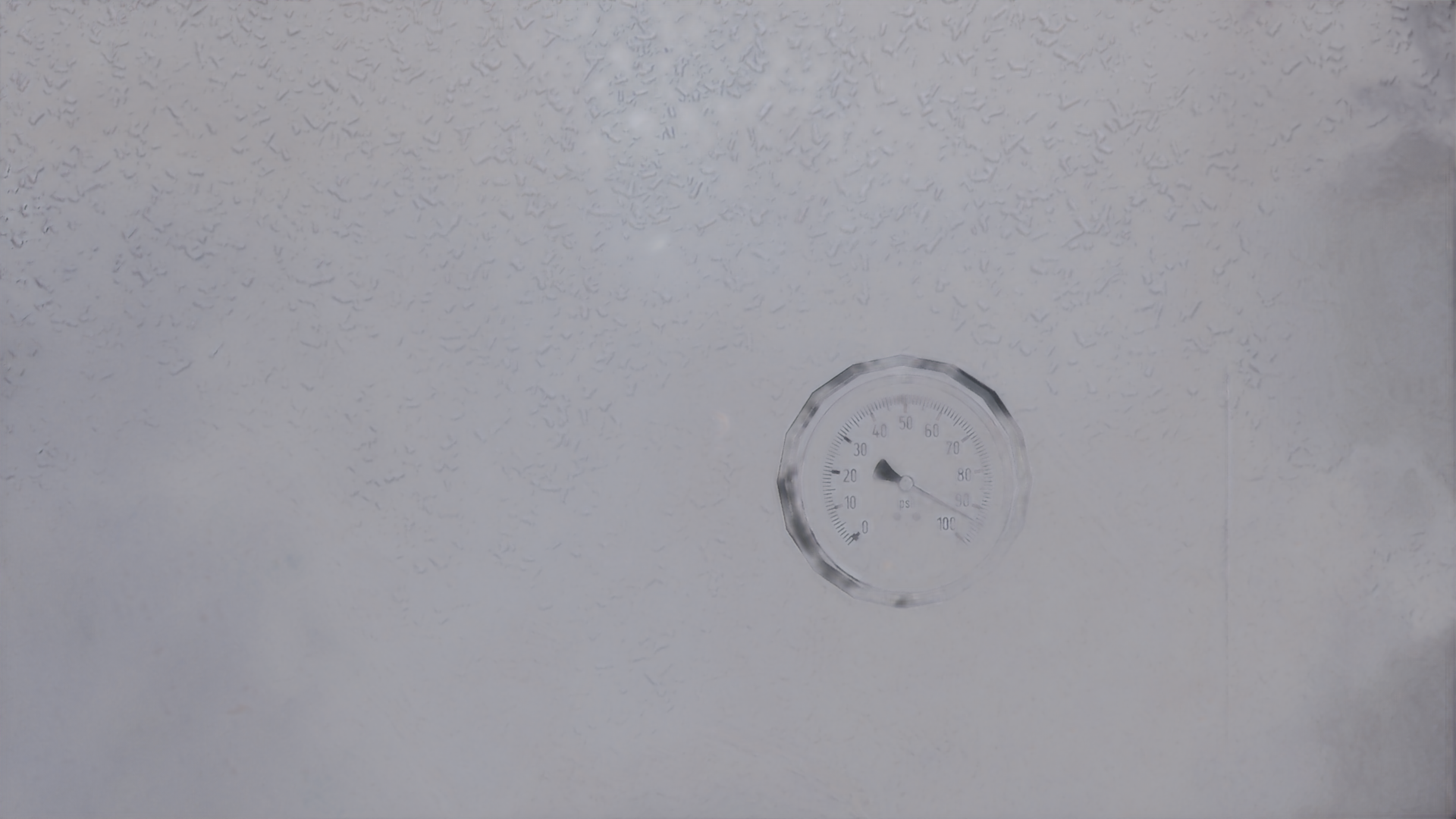}
\caption{FFA-Net 2}
\end{subtable} &
\begin{subtable}{0.22\linewidth}
\centering
\includegraphics[width=\linewidth]{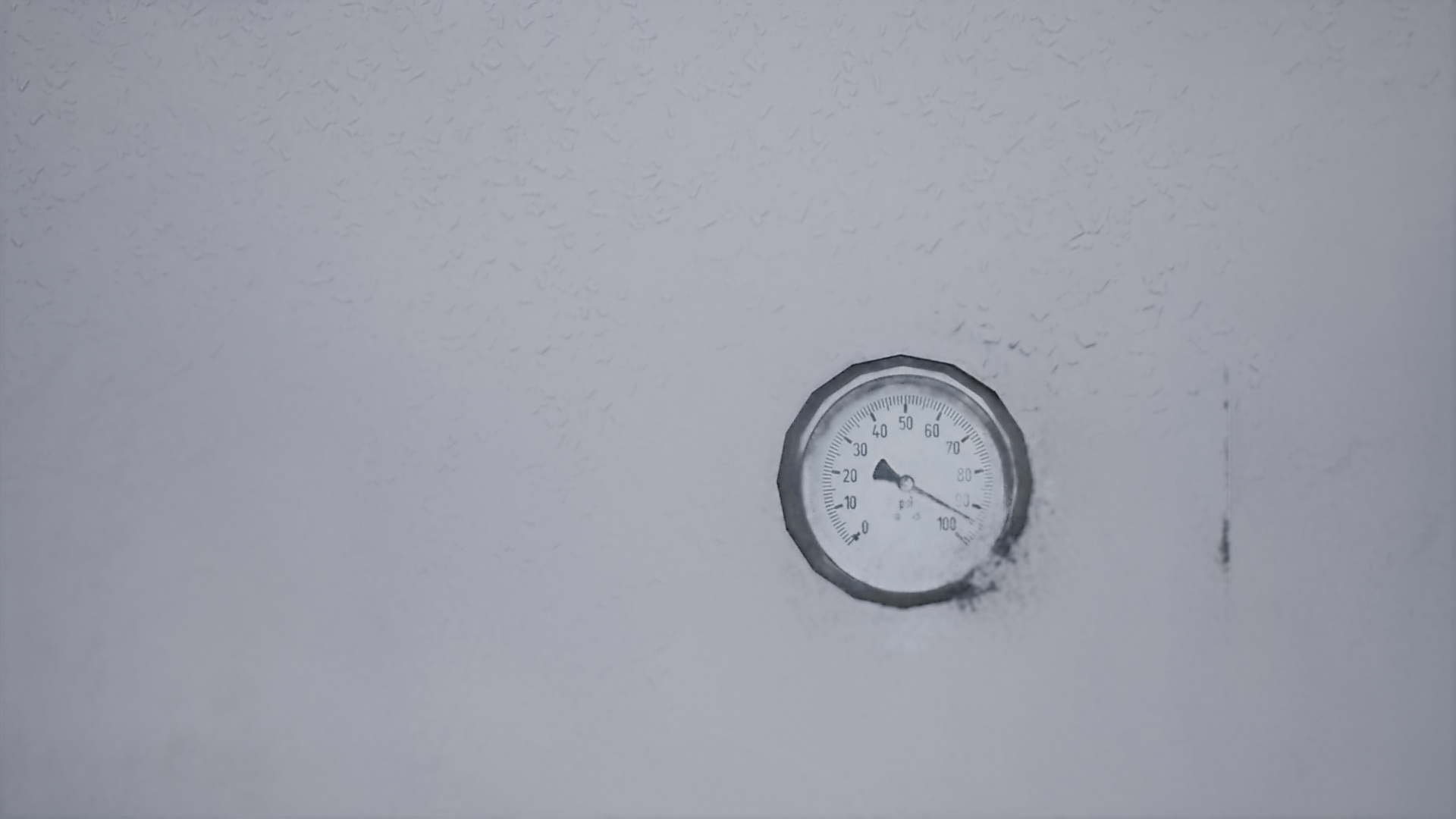}
\caption{AECR-Net 2}
\end{subtable} \\
\begin{subtable}{0.22\linewidth}
\centering
\includegraphics[width=\linewidth]{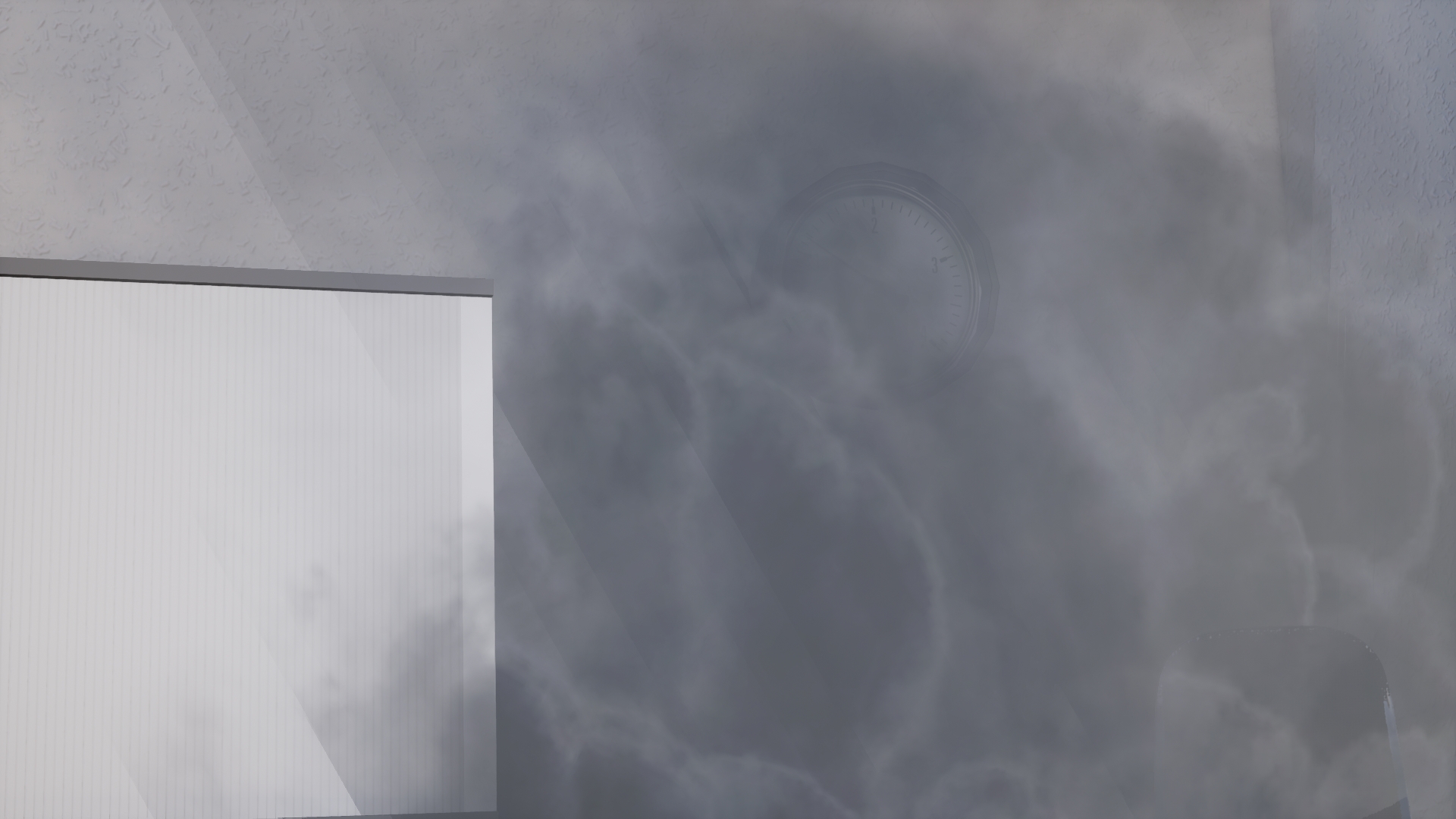}
\caption{Input 3}
\end{subtable} &
\begin{subtable}{0.22\linewidth}
\centering
\includegraphics[width=\linewidth]{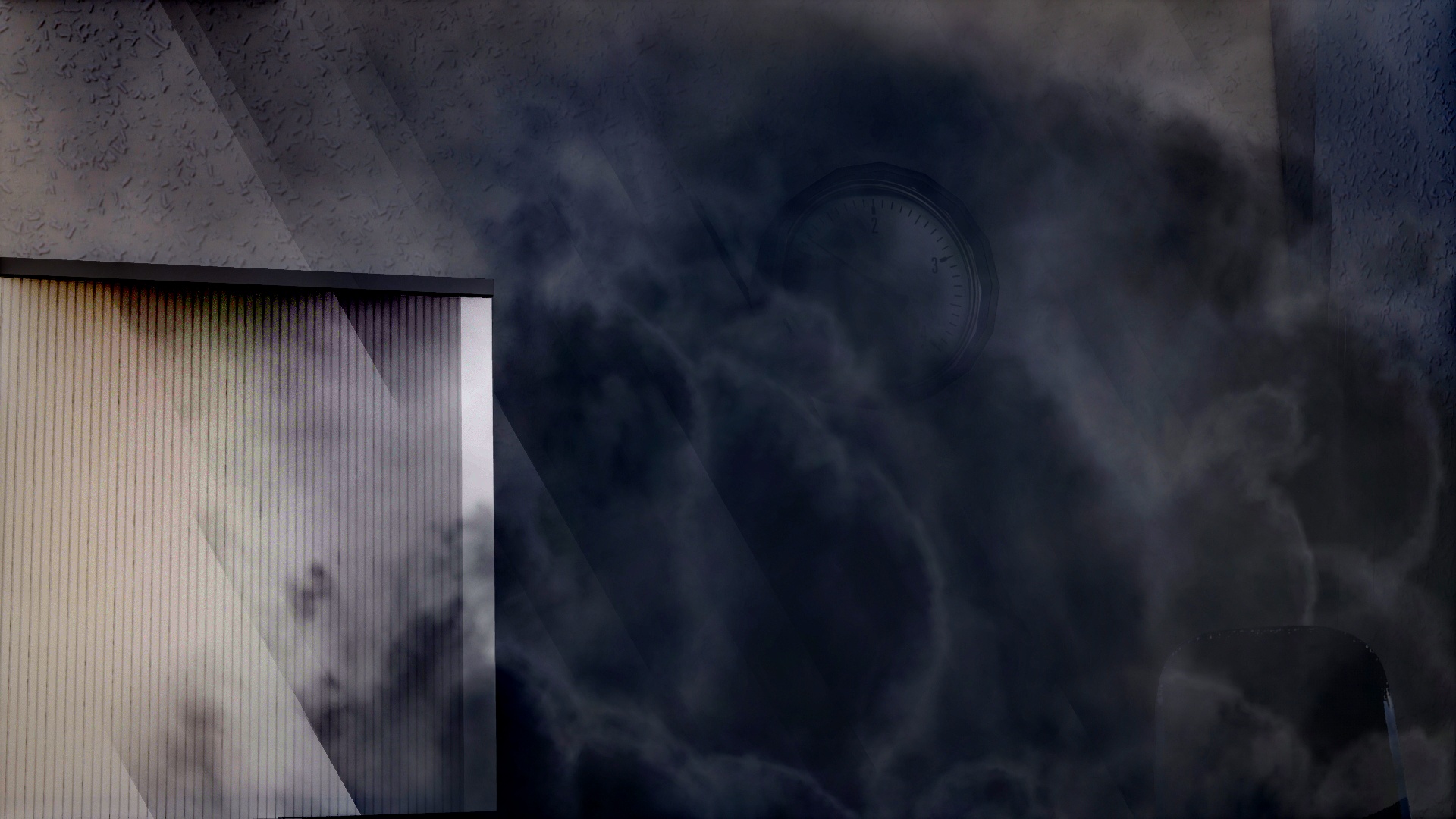}
\caption{BCCR 3}
\end{subtable} &
\begin{subtable}{0.22\linewidth}
\centering
\includegraphics[width=\linewidth]{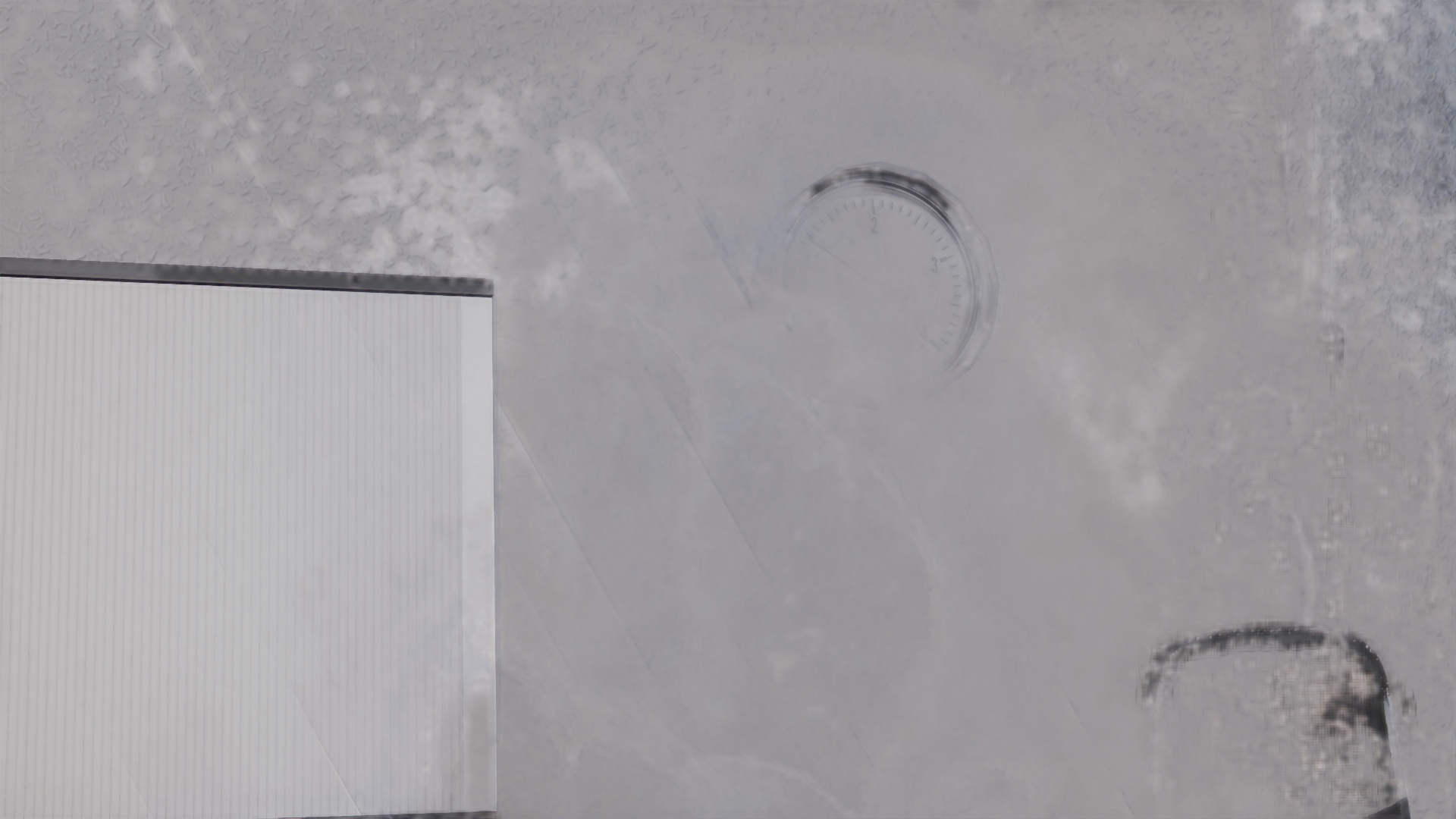}
\caption{FFA-Net 3}
\end{subtable} &
\begin{subtable}{0.22\linewidth}
\centering
\includegraphics[width=\linewidth]{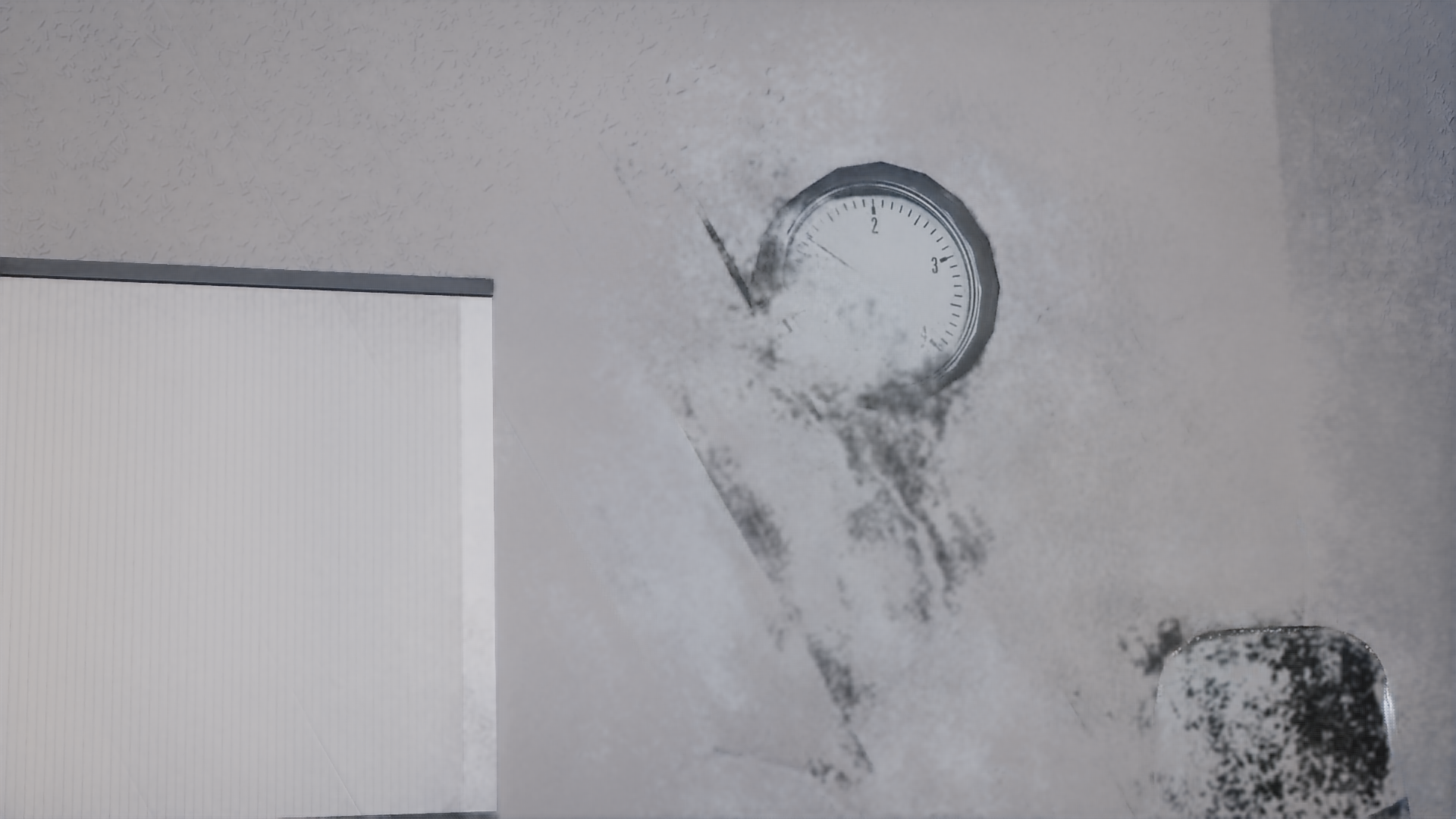}
\caption{AECR-Net 3}
\end{subtable} \\
\end{tabular}
\caption{Results of the three methods tested on three sample smoky images from \textbf{Synthetic smoke} Dataset. \textbf{First Column:} three input images with dense smoke, \textbf{second column:} output images from BCCR method, \textbf{third column:} output images from FFA-Net, \textbf{fourth column:} output images from AECR-Net.}
\label{tab:results_smoke_img}
\end{table}
%%%%%%%%%%%%%%%%%%%%%%%%%%%%%%%%%%%%%%%%%%%%%%%%%%%%%%%%%%%%%%%%%%%%%%%%%%%%%%%%%%%%%%%%%%%%%%

\section{Discussion} \label{sec:discussion}
\subsection{Comparison of the Methods}\label{sec:comparison_methods}
Both deep learning methods, when compared to the optimization method BCCR, do not attempt to estimate the scene radiance or rely on any prior. On the contrary, FFA-Net based its approach in implementing channel and pixel attention layers. As for AECR-Net it is a contrastive regularization method makes use of auto-encoders with a dynamic feature enhancement module. Both deep learning framework are an improvement compared to the prior-based method. The latter is used as the baseline benchmark algorithm for comparison purposes.\newline

\begin{table}[h!]
\centering
\begin{tabular}{|c|c|c|}
\hline
%\multicolumn{2}{|c|}{\textbf{Synthetic Haze Dataset}} \\ \hline \hline
\multicolumn{3}{|c|}{\textbf{Synthetic Haze set}} \\ \hline
\textbf{\textbf{Method}} & PSNR$_{max}$ (dB) & SSIM$_{max}$ \\ \hline
FFA-Net         &  30   & 0.96 \\ \hline
AECR-Net        &  43  & 0.98 \\ \hline
\multicolumn{3}{|c|}{\textbf{Synthetic Smoke set}} \\ \hline
%\textbf{Method} & PSNR$_{max}$ & SSIM$_{max}$ \\ \hline
FFA-Net         &  26   & 0.94 \\ \hline
AECR-Net        &  37   & 0.96 \\ \hline
\end{tabular}
\caption{Maximum PSNR and SSIM obtained during training for the FFA-Net and AECR-Net for the synthetic Haze and Smoke datasets.}
\label{tab:summary_findings}
\end{table}

\begin{figure}[h]
\centering
\includegraphics[scale=0.5]{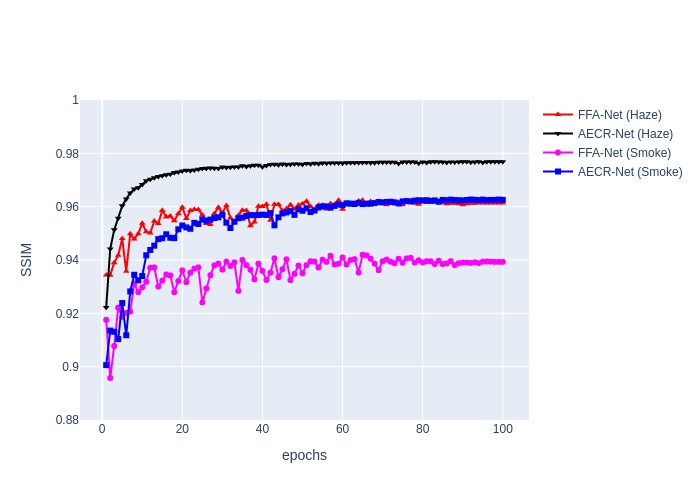}
\caption{SSIM for FFA-Net and AECR-Net.}
\label{fig:combine_ssim}
\end{figure}

\begin{figure}[h]
\centering
\includegraphics[scale=0.50]{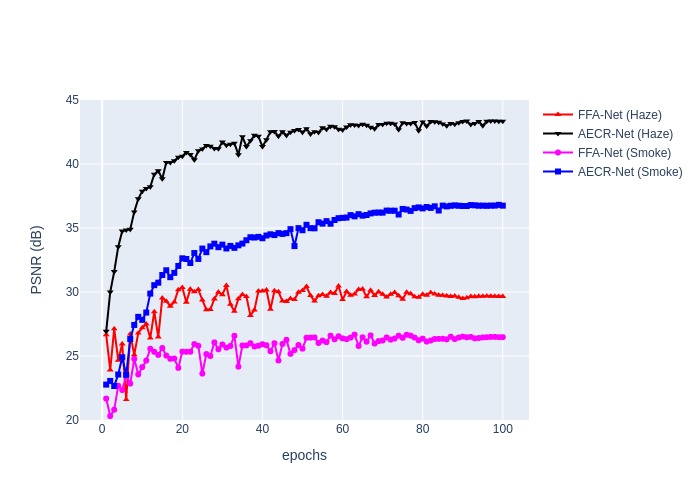}
\caption{PSNR for FFA-Net and AECR-Net.}
\label{fig:combine_psnr}
\end{figure}

Figures~\ref{fig:combine_ssim} and~\ref{fig:combine_psnr} compare the SSIM and PSNR for both datasets and deep learning methods. Table~\ref{tab:summary_findings} shows the maximun PSNR and SSIM for the FFA-Net and AECR-Net. Overall the \textit{Haze dataset} outperforms the performance with respect to the \textit{Smoke dataset}, as the scores for the \textit{Haze dataset} are higher than the \textit{Smoke dataset} counterpart, independent of the network. This result an expected outcome given that both deep learning networks FFA-Net and AECR-Net aim to remove the haze, and not to remove the smoke, from images. \newline

For benchmark datasets such as RESIDE \cite{li2019reside} the FFA-Net and AECR-Net achieve PSNR scores of 36.39\,dB and 37.17\,dB, respectively. The PSNR scores for the synthetic \textit{Haze dataset} are about 30\,dB and 43\,dB, respectively.  These results compare well with those obtained for the benchmark dataset RESIDE. For the synthetic \textit{Smoke dataset} the PSNR scores are about 26\,dB and 37\,dB for the FFA-Net and AECR-Net, respectively. Although the scores are below to the one obtained for the benchmark dataset RESIDE, the scores are of the order to those obtained by using the RESIDE dataset with architectures such as GCANet \cite{chen2018gated} and GridDehazeNet \cite{liu2019griddehazenet} (scores of the order of 30\,dB) and thus still compare well with other known networks in the literature (see also Sect.~\ref{sec:sota}). For the FFA-Net and AECR-Net the results obtained for the \textit{Haze dataset} present better performance when compared to the \textit{Smoke dataset}. In addition, adopting all results, AECR-Net displays better results when compared to FFA-Net. In general, the results obtained show the feasibility for the output models to enhance the image quality of gauges in the presence of haze or smoke.

\subsection{Sensibility of the metrics: SSIM and PSNR}\label{sec:sensibility_metrics}
%\color{red}

SSIM is designed to measure the structural similarity between two images, which may be more aligned with human visual perception. In the case of smoke and haze removal from synthetic images, if the goal is to preserve the structural integrity of the image, SSIM might be the more salient metric: It assesses the changes in structural information, making it more suitable for measuring perceived changes in complex textures and structures that are common in natural scenes. Conversely, PSNR is more related to the amount of error or noise introduced during the process. If the fidelity of the image and the degree to which the processed image deviates from the original one is of higher importance, PSNR might be of more significance. In practice, we have chosen these metrics as both should be considered in conjunction to provide a comprehensive evaluation of the model performance.\newline

In the comparative evaluation of the AECR-Net and FFA-Net machine learning models, the graphs representing PSNR and SSIM over the epochs were observed to exhibit unsmoothed characteristics due to the relatively low number of experimental runs. This limited number of runs introduces higher variance into the results, preventing a smooth trend from being established and illustrated in the respective PSNR and SSIM visualizations. In machine learning, the accuracy of a model typically does not increase linearly with the number of epochs due to the stochastic nature of the optimization algorithms such as Stochastic Gradient Descent (SGD), which can lead to substantial fluctuations in accuracy as the model navigates the high-dimensional error landscape. Additionally, the phenomenon of overfitting $-$ where the model learns the training data too well and loses generalization capability $-$ can cause a sharp decrease in accuracy after a certain number of epochs, until regularization methods take effect or the model starts learning meaningful representations again.
%\color{black}

\subsection{Autonomous Reading of Gauges}\label{sec:gr_detection}

In order to assess the effectiveness of the trained models in generating clear and enhanced images, the gauge reader detection system developed at our DLR institute~\cite{milana2022gr} was utilised (see Fig.~1 of the paper for more details). This is an end to end computer vision system that is able to autonomously read analog gauges with circular shapes and linear scales in unstructured environments. This was done to check if the output enhanced images obtained from the selected deep learning methods are actually capable of supporting post-processing applications. \newline

\begin{figure}[h!]
\centering
\includegraphics[scale=0.40]{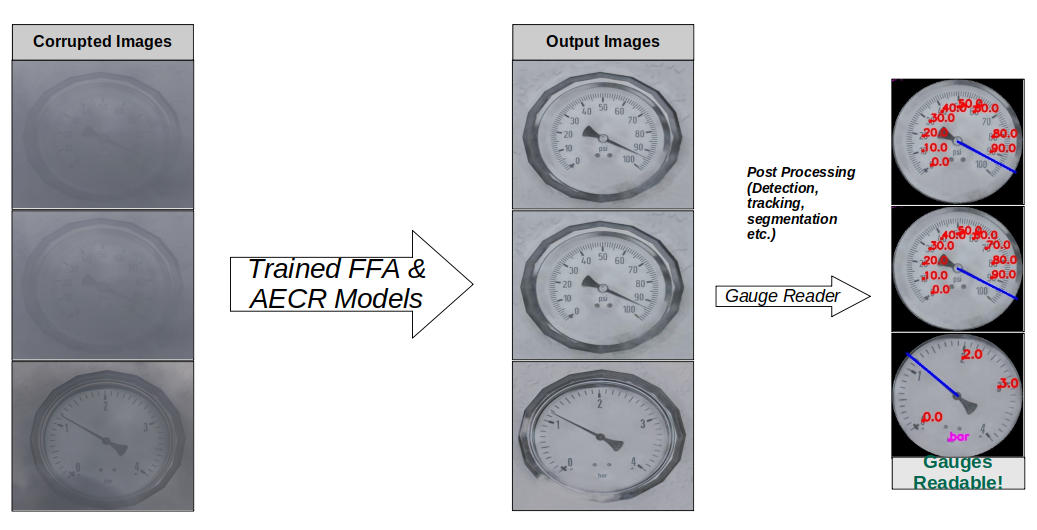}
\caption{\textbf{Left:} Analog gauge images in the presence of smoke, \textbf{Middle:} Output images obtained through trained models, \textbf{Right:} Gauge reader readings on the enhanced images obtained by using the framework of \cite{milana2022gr}.} %
\label{fig:gauge_reader_output}
\end{figure}

Fig.~\ref{fig:gauge_reader_output} shows the different steps of the post-processing. The \textit{left-panel} shows three different analog gauges in the presence of smoke. When feeding these images to the automatic gauge reader detection did not retrieve any detection. However, when the post-processed images obtained through the trained models (see \textit{middle-panel} are fed into the gauge reader detection, the gauges were not only detected but also the algorithm could accurately read the scale and needle from these enhanced images as shown in Fig.~\ref{fig:gauge_reader_output}. This demonstrates that the trained models are capable of generating clear images suitable for post-processing applications.\newline

Finally, the potential applications of these models are particularly significant in real-world scenarios where clear gauge images may not be readily available. For instance, in critical environments where contaminated images could lead to life-threatening consequences, the use of these models could prove invaluable.\newline %Furthermore, with further improvements and modifications to the FFA-Net and AECR-Net models, their effectiveness in realistic scenarios could be further enhanced.\p

\section{Conclusion} \label{sec:conclusion}
This paper acquires over 14000 images of analog gauges in the presence of synthetic haze and smoke. The datasets are fed to the deep learning models FFA-Net and AECR-Net. According to the metrics SSIM and PSNR, the architecture AECR-Net outputs a better model performance when compared to FFA-Net. Likewise, the results obtained with the synthetic Haze dataset outperforms the synthetic Smoke dataset. The results were also compared to legacy physical-based method BCCR. The deep learning frameworks delivered robust models, improving the image quality of analog gauges under affected visibility due the presence of haze or smoke. Finally, the output enhanced images of analog gauges can be successfully post-processed for automatic autonomous reading. 

\acknowledgments
O.H.R-A. acknowledges support from the DLR project \textit{Laser Assisted Metal-oxide Sensor Array} (LaMoSA) led by Kostyantin Konstantynovski at the Institute for the Protection of Terrestrial Infrastructures. O.H.R-A. and A.N.S. thank Nils Mandischer from the Institute of Mechanism Theory, Machine Dynamics and Robotics (IGMR) at RWTH Aachen University for constructive discussions. 
% References
\bibliography{report} % bibliography data in report.bib
\bibliographystyle{spiebib} % makes bibtex use spiebib.bst

\end{document}